\documentclass[a4paper,9pt, onecolumn]{article}

\usepackage{lmodern}
\usepackage[utf8]{inputenc}
\usepackage[T1]{fontenc}
\usepackage{multirow}
\usepackage{booktabs}
\usepackage[english]{babel}
\usepackage[usenames,svgnames]{xcolor}
\usepackage{setspace}
\usepackage{endnotes}
\usepackage{caption}
\usepackage{subcaption}
\usepackage{authblk}
\usepackage{ifpdf}
\ifpdf
\usepackage[pdftex]{graphicx} 
\usepackage[pdftex,pageanchor=true,hyperindex=true]{hyperref}
\hypersetup{colorlinks=true,linkcolor=black,anchorcolor=black,citecolor=black,filecolor=blue,urlcolor=blue,bookmarksnumbered=true,unicode=true,plainpages=false}
\usepackage{bookmark}
\RequirePackage{epstopdf}
\usepackage{pdfpages}
\else
\usepackage{graphicx}
\usepackage{epsfig}
\fi
\usepackage{array}
\usepackage{type1cm,soul}
\usepackage{tabularx}
\usepackage{ifthen} 
\usepackage{lscape}
\usepackage{amsmath}
\usepackage{amsfonts}
\usepackage{pifont}
\usepackage{wasysym}
\usepackage{bm}
\usepackage{fullpage}
\usepackage{xstring}
\usepackage{titlesec}
\usepackage{colortbl}
\usepackage{tcolorbox}
\usepackage{calc}
\usepackage{multirow}
\titleformat{\subsection}[runin]
{\normalfont\bfseries}{\thesubsection}{1em}{}
\titleformat{\subsubsection}[runin]
{\normalfont}{\thesubsubsection}{1em}{}
\graphicspath{{Figures/}}

\title{A semi-supervised autoencoder framework for joint generation and classification of breathing.}
\author[]{Oscar Pastor-Serrano\thanks{o.pastorserrano@tudelft.nl}}
\author[]{Danny Lathouwers, PhD}
\author[]{Zoltán Perkó, PhD\thanks{z.perko@tudelft.nl}}

\affil[]{Delft University of Technology,\\ Department of Radiation Science and Technology, Delft, Netherlands}
\date{\today}

\begin{document}
	\newcommand{\FixRef}[3][sec:]
	{\IfBeginWith{#2}{#3}
		{\StrBehind{#2}{#3}[\RefResult]}
		{\def\RefResult{#2}}\IfBeginWith{#1}{#3}
		{\StrBehind{#1}{#3}[\RefResultb]}
		{\def\RefResultb{#1}}}
	
	\newcommand{\secref}[1]
	{\FixRef{#1}{sec:}Section~\ref{sec:\RefResult}}
	\newcommand{\secreff}[1]
	{\FixRef{#1}{sec:}in Section~\ref{sec:\RefResult}}
	\newcommand{\Secreff}[1]
	{\FixRef{#1}{sec:}In Section~\ref{sec:\RefResult}}
	\newcommand{\secrefm}[2]
	{\FixRef[#2]{#1}{sec:}Sections~\ref{sec:\RefResult}-\ref{sec:\RefResultb}}
	\newcommand{\secreffm}[2]
	{\FixRef[#2]{#1}{sec:}in Sections~\ref{sec:\RefResult}-\ref{sec:\RefResultb}}
	\newcommand{\Secreffm}[2]
	{\FixRef[#2]{#1}{sec:}In Sections~\ref{sec:\RefResult}-\ref{sec:\RefResultb}}
	\newcommand{\figref}[1]
	{\FixRef{#1}{fig:}Figure~\ref{fig:\RefResult}}
	\newcommand{\figrefm}[2]
	{\FixRef[#2]{#1}{fig:}Figures~\ref{fig:\RefResult}-\ref{fig:\RefResultb}}
	\newcommand{\figreff}[1]
	{\FixRef{#1}{fig:}in Figure~\ref{fig:\RefResult}}
	\newcommand{\figreffm}[2
	]{\FixRef[#2]{#1}{fig:}in Figures~\ref{fig:\RefResult}-\ref{fig:\RefResultb}}
	\newcommand{\Figreff}[1]
	{\FixRef{#1}{fig:}In Figure~\ref{fig:\RefResult}}
	\newcommand{\Figreffm}[2]
	{\FixRef[#2]{#1}{fig:}In Figures~\ref{fig:\RefResult}-\ref{fig:\RefResultb}}
	\newcommand{\tabref}[1]
	{\FixRef{#1}{tab:}Table~\ref{tab:\RefResult}}
	\newcommand{\tabreff}[1]
	{\FixRef{#1}{tab:}in Table~\ref{tab:\RefResult}}
	\newcommand{\Tabreff}[1]
	{\FixRef{#1}{tab:}In Table~\ref{tab:\RefResult}}
	\newcommand{\tabrefm}[2]
	{\FixRef[#2]{#1}{tab:}Tables~\ref{tab:\RefResult}-\ref{tab:\RefResultb}}
	\newcommand{\tabreffm}[2]
	{\FixRef[#2]{#1}{tab:}in Tables~\ref{tab:\RefResult}-\ref{tab:\RefResultb}}
	\newcommand{\Tabreffm}[2]
	{\FixRef[#2]{#1}{tab:}In Tables~\ref{tab:\RefResult}-\ref{tab:\RefResultb}}
	\newcommand{\egyref}[1]
	{\FixRef{#1}{eq:}Equation~\ref{eq:\RefResult}}
	\newcommand{\eqreff}[1]
	{\FixRef{#1}{eq:}in Equation~\ref{eq:\RefResult}}
	\newcommand{\Eqreff}[1]
	{\FixRef{#1}{eq:}In Equation~\ref{eq:\RefResult}}
	\newcommand{\eqrefm}[2]
	{Equations~\ref{eq:#1}-\ref{eq:#2}}
	\newcommand{\eqreffm}[2]
	{\FixRef[#2]{#1}{eq:}in Equations~\ref{eq:\RefResult}-\ref{eq:\RefResultb}}
	\newcommand{\Eqreffm}[2]
	{\FixRef[#2]{#1}{eq:}In Equations~\ref{eq:\RefResult}-\ref{eq:\RefResultb}}
	\newcommand{\charef}[1]
	{\FixRef{#1}{cha:}Chapter~\ref{cha:\RefResult}}
	\newcommand{\chareff}[1]
	{\FixRef{#1}{cha:}in Chapter~\ref{cha:\RefResult}}
	\newcommand{\Chareff}[1]
	{\FixRef{#1}{cha:}In Chapter~\ref{cha:\RefResult}}
	
	\newcommand{\abs}[1]{\left|#1\right|}
	
	\def\thetable{\Roman{table}}
	\thispagestyle{empty}
	\onecolumn
	\maketitle
	\noindent

\begin{abstract}

One of the main problems with biomedical signals is the limited amount of patient-specific data and the significant amount of time needed to record the sufficient number of samples needed for diagnostic and treatment purposes. In this study, we present a framework to simultaneously generate and classify  biomedical time series based on a modified Adversarial Autoencoder (AAE) algorithm and one-dimensional convolutions. Our work is based on breathing time series, with specific motivation to capture breathing motion during radiotherapy lung cancer treatments. First, we explore the potential in using the Variational Autoencoder (VAE) and AAE algorithms to model breathing from individual patients. We extend the AAE algorithm to allow joint semi-supervised classification and generation of different types of signals. To simplify the modeling task, we introduce a pre-processing and post-processing compressing algorithm that transforms the multi-dimensional time series into vectors containing time and position values, which are transformed back into time series through an additional neural network. By incorporating few labeled samples during training, our model outperforms other purely discriminative networks in classifying breathing baseline shift irregularities from a dataset completely different from the training set. To our knowledge, the presented framework is the first approach that unifies generation and classification within a single model for this type of biomedical data, enabling both computer aided diagnosis and augmentation of labeled samples within a single framework.
\end{abstract}

\section{Introduction}
Biomedical data is the driving force behind most modern advances in medicine. The use of biomedical records is associated however with a series of problems such as the lack of reliable models capable of simulating data with clinical precision, the absence of personalized models for diagnosis, or the lack of labeled samples since the labels containing personal features that compromise privacy or simply are not recorded \cite{intro1}. Some of the initial efforts to model biomedical data include analytical approaches: e.g., McSharry et al. \cite{ecg1} developed an electrocardiogram (ECG) model based on three coupled ordinary differential equations, and George et al. \cite{br1} introduced a sinusoidal model to represent breathing.

Recent advances in Deep Learning and the introduction of algorithms such as the Variational Autoencoder (VAE) \cite{vae}\cite{vae2} and Generative Adversarial Networks (GANs) \cite{gan} have resulted in a wide variety of methods capable of generating and classifying biomedical signals, most of them having been applied to ECG data. Regarding classification, Acharya et al. \cite{c1} \cite{c2}, Fujita et al. \cite{c3} and Cimr et al. \cite{c4} present classification Convolutional Neural Network (CNN) frameworks for computer aided diagnosis based on biomedical signals. Yildirim et al. \cite{gen0} propose an efficient algorithm based on autoencoder artificial neural networks (ANNs) that compresses ECG signals but lacks generative capabilities. Recent implementations of CNN architectures result in minimal classification error of ECG signals \cite{c5} \cite{c6}. With respect to generation, both Zhu et al. \cite{gen1} and Delany et al. \cite{gen2} propose generative models of realistic ECG signals that combine different ANN architectures (recurrent and convolutional) under a GAN adversarial training objective. Golany and Radinsky \cite{gen3} present a framework where a GAN generates data for ECG classification, while Wulan et al. \cite{gen4} introduce an autoregressive model able to produce longer signals with high variability.

Most of the previously proposed methods focus either on generation or classification and result in models that depend on large labeled datasets and supervised training; are resource intensive and require significant amounts of computing power; are inaccurate when the dataset is imbalanced (there are very few labels for some classes of interest), or generate data that lacks variability and has a limited temporal dependence \cite{ov1} \cite{ov2}. Furthermore, most of the approaches are not capable of capturing the structure of the data in a low-dimensional manifold in which specific regions correspond to similar samples.

In this study, we focus on mechanical breathing signals representing the movement of chest markers during respiration.  Among their many applications, these type of biomedical signals are of great importance in radiotherapy cancer treatments, where they are used to quantify the impact of respiration and to design robust lung cancer radiotherapy treatments that withstand the detrimental effect of breathing motion during treatment delivery. Among the most important breathing irregularities are baseline shifts, which are gradual or sudden changes in the exhale position and trend of respiration. Baseline shifts negatively affect the outcome of radiotherapy treatments \cite{bs1}. To our knowledge, there are no previous studies that develop breathing generative models that result in realistic respiratory traces. Likewise, very few computer-aided diagnostic tools have been presented for physical breathing signals. Abreu et al. \cite{bs2} present an autoencoder framework that discriminates between apnea and regular breathing, focusing on gating radiotherapy treatments.  

We investigate whether it is possible to combine classification and generation of breathing signals within a single model. We propose \textbf{a semi-supervised framework that simultaneously classifies and generates breathing motion with high accuracy using a small subset of labeled data}, and which \textbf{outperforms purely discriminative models}, and could in principle be applied to modeling other biomedical signals. The main contributions of this research are threefold. First, we investigate the suitability of probabilistic generative models based on one-dimensional convolutional filters for the task of modeling breathing signals. Second, building upon these breathing models, we introduce a modified semi-supervised algorithm to train a joint generative-discriminative model using a partially-labeled dataset. The proposed model can be used to simultaneously generate and classify samples of irregular breathing or samples from a population of patients. Third, we develop a pre-processing and post-processing method that transforms back and forth the breathing signals from their original 3-dimensional time series form into a simplified vector form containing pairs of position-time values. This transformation significantly reduces the dimensionality of the inputs and speeds up training. 

\section{Background}
\subsection*{Probabilistic generative models.} Consider $\mathrm{\mathbf{x}}\in\mathbb{R}^M$ to be a random vector over a vector space $\mathcal{X}$, with unknown underlying probability distribution $p_{data}(\mathrm{\mathbf{x}})$. Given a dataset  $\mathcal{D}=\{\bm{x}^{(i)}\}_{i=1}^{N_{\mathcal{D}}}$ with $N_{\mathcal{D}}$ independent and identically distributed (i.i.d) data points, the goal is to model a probability distribution $p_{\bm{\theta}}(\bm{\mathrm{\mathbf{x}}})$ that approximates the unknown true probability distribution generating the data using a probabilistic graphical model with parameters $\bm{\theta}$. Let this probabilistic model be a latent variable model, which conditions the observed variable $\mathrm{\mathbf{x}}$ on the unobserved random variable $\mathrm{\mathbf{z}}\in\mathbb{R}^N$ over the latent space $\mathcal{Z}$ containing $N$ latent variables that are assumed to capture the principal factors of variation in the data. The latent variable model represents the joint distribution of observed and unobserved variables and factorizes as $p_{\bm{\theta}}(\mathrm{\mathbf{x}},\mathrm{\mathbf{z}})=p_{\bm{\theta}}(\mathrm{\mathbf{x}}|\mathrm{\mathbf{z}})p(\mathrm{\mathbf{z}})$. The (target) marginal distribution of the observed variables can be recovered as

\begin{equation}
	p_{\bm{\theta}}(\bm{x}) = \int_\mathcal{Z} p_{\bm{\theta}}(\bm{x},\bm{z}) d\bm{z} = \int_\mathcal{Z}  p_{\bm{\theta}}(\bm{x}|\bm{z})p(\bm{z})d\bm{z},
	\label{eq:marg}
\end{equation}  

\noindent where $p(\bm{z})$ is the prior probability distribution over $\mathcal{Z}$ and $p_{\bm{\theta}}(\bm{x}|\bm{z})$ is a conditional distribution that can be parametrized using neural networks. In principle, the prior could be any function and it is not conditioned on the observations. Point-estimates of the parameters $\bm{\theta}$ of the latent variable model can be obtained via maximum likelihood estimation, i.e., by maximizing  the (log-) marginal distribution of the observed data

\begin{equation}
\bm{\theta}^* = \underset{\bm{\theta}}{\text{argmax}} \: \sum_{\bm{x}\in\mathcal{D}}\log\:(p_{\bm{\theta}}(\bm{x})) \simeq \underset{\bm{\theta}}{\text{argmax}} \:\: \mathbb{E}_{\mathrm{\mathbf{x}}\sim\hat{p}_{data}(\bm{x})}\log (p_{\bm{\theta}}(\bm{x})),
\label{eq:opt}
\end{equation}

\noindent where the expected value is computed over the empirical data distribution $\hat{p}_{data}(\bm{x})$. The empirical data distribution is different from the true underlying data generating distribution $p_{data}(\bm{x})$ to which we do not have direct access and we want to approximate. $\hat{p}_{data}(\bm{x})$ is defined as a mixture of Dirac delta distributions $\delta(\bm{x})$ that assigns probability mass $1/N_{\mathcal{D}}$ to each data point in $\mathcal{D}$ as

\begin{equation}
\hat{p}_{data}(\bm{x})=\frac{1}{N_{\mathcal{D}}}\sum_{i=1}^{N_{\mathcal{D}}}\delta(\bm{x}-\bm{x}^{(i)}).
\end{equation}

In practice, computing the integral over the space $\mathcal{Z}$ in \egyref{eq:marg} is intractable. Thus, the optimization in \egyref{opt} is simplified by maximizing a lower bound on the marginal distribution.

\subsection*{Variational Autoencoder.} \label{sec:vae}  Kingma and Welling \cite{vae}, and Rezende et al. \cite{vae2} present an algorithm that allows to estimate the latent variable model parameters maximizing the Evidence Lower BOund (ELBO). The algorithm, known as Variational Autoencoder (VAE), requires an inference model that approximates the (also) intractable true posterior distribution $p_{\bm{\theta}}(\bm{z}|\bm{x})$ using a family of probability distributions of the latent variables $q_{\bm{\phi}}(\bm{z}|\bm{x})$ conditioned on observed data points, with parameters $\bm{\phi}$ shared across data points $\bm{x}$. By including the inference model, the ELBO optimization objective is formulated as

\begin{equation}
\log\:(p_{\bm{\theta}}(\bm{x})) \geq\mathbb{E}_{\mathrm{\mathbf{z}}\sim q_{\bm{\phi}}(\bm{z}|\bm{x})}[\log(p_{\bm{\theta}}(\bm{x}|\bm{z}))] - D_{KL}(q_{\bm{\phi}}(\bm{z}|\bm{x})||p(\bm{z})):=\text{ELBO}(\bm{\theta},\bm{\phi},\bm{x}),
\end{equation}

\noindent where the second term is the Kullback - Leibler (KL) divergence, denoted $D_{KL}(\cdot ||\cdot)$. Essentially, the KL divergence quantifies "the difference" between distributions. Further details about the ELBO and how to compute the KL-divergence are included in Appendix \ref{app:elbo}.

In the VAE framework, the prior is the multivariate Gaussian $p(\bm{z})=\mathcal{N}(\bm{z};\bm{0},\bm{I})$, where $\bm{I}$ is the identity matrix. The likelihood conditional distribution $p_{\bm{\theta}}(\bm{x}|\bm{z})$ is represented as a multivariate Gaussian probability distribution with identity covariance matrix $p_{\bm{\theta}}(\bm{x}|\bm{z}) = \mathcal{N}(\bm{x};f_{\bm{\theta}}(\bm{z}),\bm{I})$, where the function $f_{\bm{\theta}}(\bm{z}):\mathcal{Z}\rightarrow\mathbb{R}^M$ is parameterized with an ANN referred to as the probabilistic decoder. With this formulation,  $p_{\bm{\theta}}(\bm{x}) $ is an infinite mixture of Gaussian distributions. In the same way as with the probabilistic decoder, it is possible to parameterize the inference model conditional distribution using a neural network that performs a mapping $g_{\bm{\phi}}(\bm{x}):\bm{x}\in\mathcal{X}\rightarrow (\bm{\mu}(\bm{x}),\bm{\sigma}(\bm{x}))\mathbb{R}^{2N}$ and outputs the mean $\bm{\mu}(\bm{x})$ and standard deviation $\bm{\sigma}(\bm{x})$ of the Gaussian distribution $q_{\bm{\phi}}(\bm{z}|\bm{x}) = \mathcal{N}(\bm{z};\bm{\mu}(\bm{x}), \text{diag}\:\bm{\sigma^2(\bm{x})}).$ 

The ELBO balances two terms: the first term encourages the probabilistic decoder to produce samples that resemble the observed data, while the second term forces the approximated posterior distribution obtained from the inference model to be close to the prior distribution. Using the negative ELBO as optimization objective, the minimization problem to solve is:

\begin{equation}
\bm{\theta}^*,\bm{\phi}^* = \underset{\bm{\theta},\bm{\phi}}{\text{argmin}}\:\mathbb{E}_{\mathrm{\mathbf{x}}\sim\hat{p}_{data}(\bm{x})} \big[ -\mathbb{E}_{\mathrm{\mathbf{z}}\sim q_{\bm{\phi}}(\bm{z}|\bm{x})}[\log(p_{\bm{\theta}}(\bm{x}|\bm{z}))] + \beta D_{KL}(q_{\bm{\phi}}(\bm{z}|\bm{x})||p(\bm{z}))\big],
\label{eq:elbo}
\end{equation}

\noindent where $\beta$ is a hyperparameter that can be used to weigh the reconstruction and regularization terms \cite{betavae}. The minimization in \egyref{elbo} can be performed using first order stochastic methods such as Stochastic Gradient Descent (SGD). The reparametrization trick is usually employed to propagate the gradients of the weights through the encoder, as described in \cite{vae}. Details on the VAE algorithm and how to estimate its gradients can be found in \cite{vae}\cite{vaei}.

\subsection*{Adversarial Autoencoder.}\label{sec:aae} Makhzani et al. \cite{aae} propose an alternative formulation to the ELBO, where the KL divergence is approximated as the optimal value of an adversarial loss that forces the aggregated posterior distribution $q_{\bm{\phi}}(\bm{z})$ to be close to the prior:

\begin{equation}
q_{\bm{\phi}}(\bm{z}) = \int_{\mathcal{X}} q_{\bm{\phi}}(\bm{z}|{\bm{x}}) \hat{p}_{data}(\bm{x}) d\bm{x}\:\: \simeq p(\bm{z}).
\end{equation}

In the original paper, the authors explore the use of both probabilistic encoders and deterministic encoders with $g_{\bm{\phi}}(\bm{x})$ as a deterministic mapping. We use a universal approximation probabilistic encoder that in principle is able to learn any arbitrary posterior distribution by employing random noise $\eta\in H\in\mathbb{R}$ with distribution $p(\eta)=\mathcal{N}(\eta;0,1)$. Such encoders take additional random noise values to produce samples $\bm{z}=g_{\bm{\phi}}(\bm{x},\eta)$, and can use different noise values $\eta$ to map the same input $\bm{x}$ to a domain in $\mathcal{Z}$. The aggregated posterior can be computed as

\begin{equation}
q_{\bm{\phi}}(\bm{z}) = \int_{\mathcal{X}}\int_{H} \delta(\bm{z}-g_{\bm{\phi}}(\bm{x},\eta)) p(\eta) \hat{p}_{data}(\bm{x}) d\eta d\bm{x},
\end{equation}

The adversarial loss is based on GANs. Let the encoder network be $g_{\bm{\phi}}(\bm{x},\eta)$ with parameters $\bm{\phi}$ that performs a mapping $g_{\bm{\phi}}(\bm{x},\eta):\mathcal{X}\times H\rightarrow\mathcal{Z}$. A discriminator model is introduced, modeled also with an ANN with mapping function $d_{\bm{\xi}}(\bm{z}):\mathcal{Z}\rightarrow\mathbb{R}$ that outputs a single scalar logit. The value $S(d_{\bm{\xi}}(\bm{z}))\in[0,1]$ represents the probability that $\bm{z}$ is a sample from the prior distribution $p(\bm{z})$ (true samples) rather than being a latent space mapping from the encoder (fake samples), where $S(z):=(1+e^{-z})^{-1}$ is the logistic sigmoid function. This translates into a min-max optimization problem

\begin{equation}
	\underset{\bm{\phi}}{\min}\:\:\underset{\bm{\xi}}{\max}\:\:\mathbb{E}_{\mathrm{\mathbf{z}}\sim p(\bm{z})}[\log (S(d_{\bm{\xi}}(\bm{z})))]+\mathbb{E}_{\mathrm{\mathbf{x}}\sim \hat{p}_{data}(\bm{x})}\mathbb{E}_{\mathrm{\eta}\sim p_{(\eta)}}[\log (1-S(d_{\bm{\xi}}(g_{\bm{\phi}}(\bm{x},\eta))))],
	\label{eq:aaeo}
\end{equation}

\noindent where first the discriminator is trained to correctly distinguish between real and encoder samples by maximizing the probability of classifying real samples from the prior $\bm{z}_r$ as real ($S(d_{\bm{\xi}}(\bm{z}_r)=1$)) and fake samples from the encoder $\bm{z}_f$ as false ($S(d_{\bm{\xi}}(\bm{z}_f)=0$)). Second, the encoder is trained to minimize the probability  $1-S(d_{\bm{\xi}}(\bm{z}_f))$ that the discriminator identifies its samples $\bm{z}_f$ as fake, where $d_{\bm{\xi}}(\bm{z_f})=1$ means that the discriminator classifies a fake sample as a true sample. Training the probabilistic decoder $p_{\bm{\theta}}(\bm{x}|\bm{z})$, the inference model $q_{\bm{\phi}}(\bm{z}|{\bm{x}})$ and the discriminator $d_{\bm{\xi}}(\bm{z}_f)$ can be done with SGD in two alternating steps: a reconstruction phase forces the decoder to produce realistic samples by using the $\bm{z}_f$ variables produced by the inference model, and the regularization phase updating the parameters of the encoder and discriminator. As shown in Appendix \ref{app:aaeloss}, optimizing the adversarial objective results in an approximation to the ELBO, where the optimum discriminator function is $d_{\bm{\xi}}^* = \log(q(z)/p(z))$ and the regularization term  $\mathbb{E}_{\mathrm{\mathbf{x}}\sim \hat{p}_{data}(\bm{x})}[D_{KL}(q_{\bm{\phi}}(\bm{z}|\bm{x})||p(\bm{z}))]$ in \egyref{elbo} is replaced by $D_{KL}(q_{\bm{\phi}}(\bm{z})||p(\bm{z}))$.  More details about the adversarial objective can be found in Appendix \ref{app:aaeloss}. 

\section{Joint generative-discriminative models.} One of the advantages of the AAE algorithm is that the standard architecture can be slightly modified in order to additionally perform semi-supervised classification based on few labeled data points. The most notable difference with respect to the standard AAE architecture in \cite{aae} is the introduction of an extra discrete latent variable $\bm{y}\in\{0,1\}^C$, which represents the class to which the input belongs over $C$ classes. The class $\bm{y}$ is practically implemented as a sparse one-hot vector with a 1 entry at the position corresponding to the class. In the case of breathing, the $\bm{y}$ variable could indicate the presence of irregularities or the patient to which breathing pertains, while for ECG $\bm{y}$ could represent type of heart arrhythmia. The encoder now outputs the joint distribution $q_{\bm{\phi}}(\bm{y},\bm{z}|\bm{x})$ that factorizes as

\begin{equation}
	q_{\bm{\phi}}(\bm{y},\bm{z}|\bm{x}) = q_{\bm{\phi}}^c(\bm{y}|\bm{x})q_{\bm{\phi}}^s(\bm{z}|\bm{x}),
\end{equation}

\noindent where $q_{\bm{\phi}}^c(\bm{y}|\bm{x})$ is a distribution that performs a mapping  $\bm{\pi}(\bm{x},\eta):\mathcal{X}\times H\rightarrow\mathbb{R}^C$ based on the input $\bm{x}$ and noise $\eta$. The use of the softmax non-linearity and the use of one-hot vectors as a target forces sparsity in $\bm{y}=\bm{\pi}(\bm{x},\eta)$. The approximate posterior $q_{\bm{\phi}}^s(\bm{z}|\bm{x})$ is either a distribution or a deterministic mapping, as in the standard AAE. In the original paper \cite{aae}, the semi-supervised AAE is trained to perform either clustering or generation. Given that our goal is to simultaneously classify and generate new samples given a specific input, we propose a modified AAE architecture that uses a single discriminator for both the classification and style heads. In this way, the aggregated approximated posterior is forced to match the mixture prior distribution

\begin{equation}
	q_{\bm{\phi}}(\bm{z},\bm{y}) = \int_{\mathcal{X}} q_{\bm{\phi}}^s(\bm{z}|{\bm{x}})q_{\bm{\phi}}^c(\bm{y}|{\bm{x}}) \hat{p}_{data}(\bm{x}) d\bm{x}\:\: \simeq p(\bm{z},\bm{y}).
\end{equation}

\noindent where the prior distribution factorizes as the mixture

$$p(\bm{z},\bm{y}) = p(\bm{z}) p(\bm{y})= \mathcal{N}(\bm{z};\bm{0},\bm{I}) \text{Cat}(\bm{y};\bm{c}) ,$$

With this setup, each label $\bm{y}$ is associated with an independent low-dimensional space where $\bm{z}$ is distributed according to $p(\bm{z})$. Sampling from each cluster is easy, as opposed to the models presented in \cite{aae} that are specifically trained either for clustering or conditional generation of samples, and where $\bm{z}$ is jointly distributed according to $p(\bm{z})$ over all $\bm{y}$ classes.

\subsection*{Semi-supervised models.}  Let $\hat{p}_{data}(\bm{x}_l,\bm{y}_l)$ be the joint empirical distribution of labeled data $\bm{x}_l$ with labels $\bm{y}_l$. Our variant of the AAE, named Semi-supervised AAE (SAAE) in the remainder of the paper, is trained in 3 stages: a reconstruction and regularization phase that are identical to the ones in the standard AAE, and a supervised classification phase for the available labels in which the cross-entropy $\alpha\cdot\mathbb{E}_{\mathrm{\mathbf{x}_l},\mathrm{\mathbf{y}_l}\sim \hat{p}_{data}(\bm{x}_l,\bm{y}_l)}[-\log q_{\bm{\phi}}^c(\bm{y_l}|{\bm{x_l}})]$ is minimized, where $\alpha$ controls the weight of the classification loss. The optimization problem is defined as

\begin{flalign}
	\text{\footnotesize Regularization:}\:\: \underset{\bm{\xi}}{\text{max}} \:\mathbb{E}_{\mathrm{\mathbf{z}},\mathrm{\mathbf{y}}\sim p(\bm{z},\bm{y})}[\log (S(d_{\bm{\xi}}(\bm{z},\bm{y})))]+\mathbb{E}_{\mathrm{\mathbf{x}}\sim \hat{p}_{data}(\bm{x})}\mathbb{E}_{\mathrm{\eta}\sim p_{(\eta)}}[\log (1-S(d_{\bm{\xi}}(g_{\bm{\phi}}(\bm{x},\eta))))] &&
\end{flalign}
\begin{flalign}
	\text{\footnotesize Classification:}\:\:\:\:\: \underset{\bm{\phi}}{\text{min}}\: \alpha\cdot\mathbb{E}_{\mathrm{\mathbf{x}_l},\mathrm{\mathbf{y}_l}\sim \hat{p}_{data}(\bm{x}_l,\bm{y}_l)}[-\log q_{\bm{\phi}}^c(\bm{y}_l|{\bm{x}_l})] && 
\end{flalign}
\begin{flalign}
	\text{\footnotesize Reconstruction:}\:\: \underset{\bm{\theta},\bm{\phi}}{\text{max}}\: \mathbb{E}_{\mathrm{\mathbf{x}}\sim\hat{p}_{data}(\bm{x})} \mathbb{E}_{\mathrm{\mathbf{z},\mathbf{y}}\sim q_{\bm{\phi}}(\bm{z},\bm{y}|\bm{x})}[\log(p_{\bm{\theta}}(\bm{x}|\bm{z},\bm{y}))+\mathbb{E}_{\mathrm{\mathbf{x}}\sim \hat{p}_{data}(\bm{x})}\mathbb{E}_{\mathrm{\eta}\sim p_{(\eta)}}[d_{\bm{\xi}}(g_{\bm{\phi}}(\bm{x},\eta))]&& 
\end{flalign}

\section{Methods and materials}

\begin{figure}[]
	\centering
	\includegraphics[width=\textwidth]{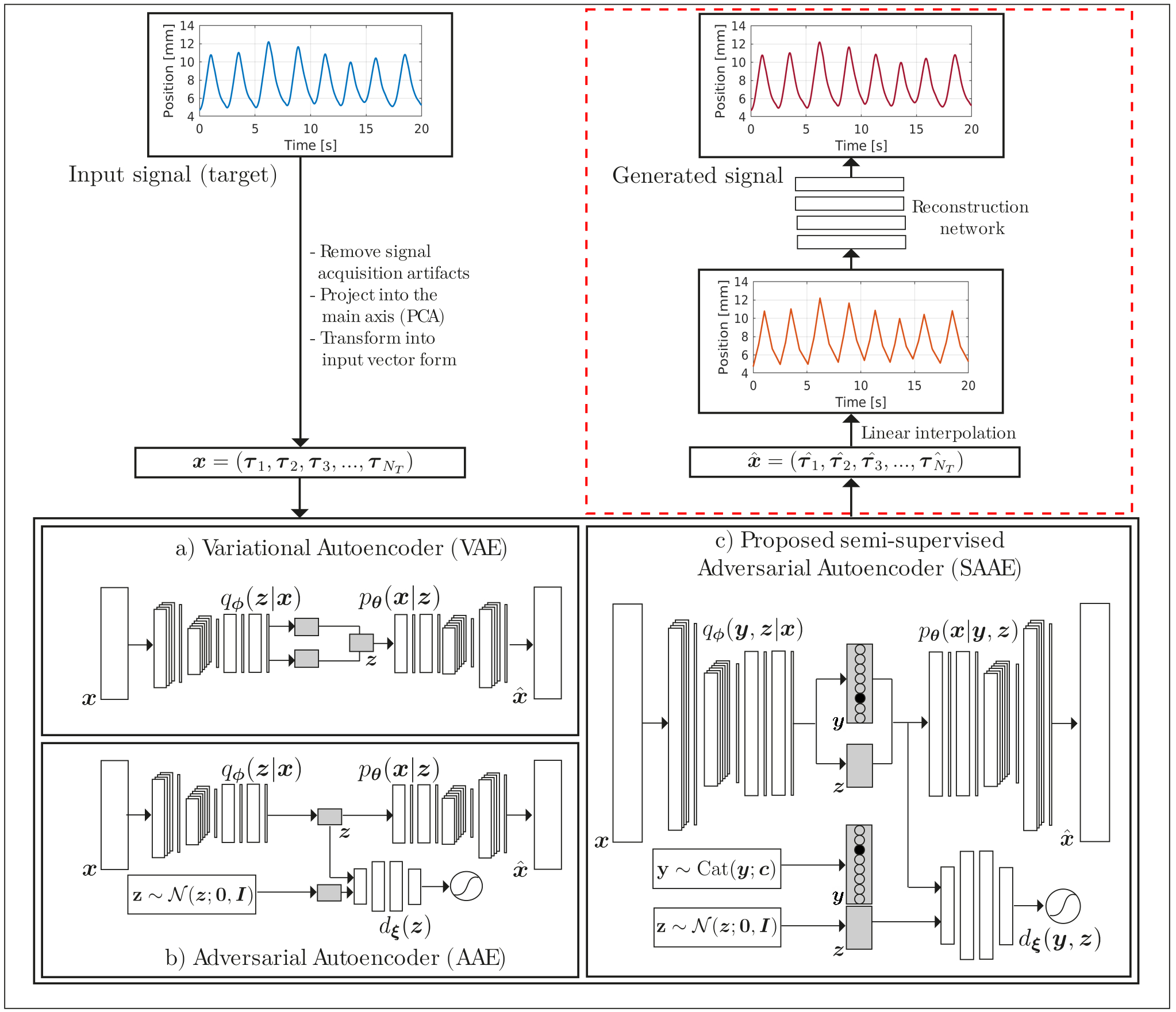} 
	\caption{Summary of the breathing modeling workflow. First, the original time series is pre-processed and projected into the main axis of movement (eigenvector with biggest eigenvalue) using Principal Component Analysis (PCA), from which the input vectors $\bm{x}$ are obtained. Patient or population models are then obtained through the use of the (a) VAE, (b) AAE and (c) SAAE with one-dimensional convolutional encoder and decoder models. In the VAE and AAE, the encoder (or inference) model produces a low-dimensional latent variable $\bm{z}$ that ideally captures the factors of variation in the dataset, such as variations in period, amplitude and exhale position. In the SAAE the inference model generates a class label latent variable $\bm{y}$ besides vector $\bm{z}$. Labeled data can be leveraged during training in order to learn the classification task in a semi-supervised manner. During generation (red dashed square), the sampled latent variables are transformed into the input vector form. These new vectors $\hat{\bm{x}}$ are then transformed into a time series with the help of an auxiliary reconstruction neural network.}
	\label{fig:ssaae}
\end{figure}

First, we investigate the benefits of applying VAE and the AAE to model respiratory motion of individual patients using few latent parameters. Second, using the presented SAAE architecture, we obtain a population breathing model capable of simultaneously classifying and generating specific types of breathing. We base our study on breathing signals, which are time series representing the position of chest markers in lung cancer patients. \figref{ssaae} shows an overview of the workflow, including the pre-processing, the models for classification and generations, and the final post-processing time series reconstruction step. 

\subsection*{Patient and population data.} Different breathing signals were obtained with the stereotactic radiosurgery system Cyberknife\textsuperscript{\tiny\textregistered} (Accuray Inc., Sunnyvale CA, US). Cyberknife\textsuperscript{\tiny\textregistered} tracks breathing movement using correspondence of markers positioned on the patient's chest \cite{ck}. The data used in our study consists of long respiratory traces for 21 different patients. The optical device tracks data with a 26 Hz frequency, for a total duration between ten and thirty minutes. The breathing signals for 15 out of the 21 patients were obtained from the open-access database recorded at Georgetown University Hospital (Washington D.C, United States) \cite{sig}, with breathing amplitudes in the interval (0.5,10) mm. The 6 remaining respiratory traces were recorded during treatments at Erasmus MC (Rotterdam, Netherlands) and correspond to 6 patients with much smaller amplitudes in the range (0.5,2) mm. The 2 datasets are referred to as the GUH and EMC datasets for the remainder of the paper.

\subsection*{Input data \& pre-processing.} The first step consists of removing obvious errors in the signal acquisition process that are usually related to machine recalibration during measurement. This results in a 3D time series, where each dimension correspond to a physical dimension in the Cartesian coordinate system. Since the 3D are correlated, the 3D signals are further compressed into a 1D signal by using Principal Component Analysis (PCA) and projecting them onto the main axis of movement, which is the eigenvector with highest eigenvalue. We find that the projection onto the principal axis retains around 95\% of the original variance. The resulting trace is divided into different periods $\bm{\tau}_j$, each of them corresponding to the time between start of different inhales. Each period $j$ is discretized into 4 points with $A_{s,j}$ denoting position and a $\Delta_{s,j}$ representing the difference in time between consecutive points. Thus, a period is parametrized by the vector

\begin{equation}
\bm{\tau}_j = (A_{\text{EE},j}, \Delta_{\text{EE},j}, A_{\text{MI},j}, A_{\text{EI},j}, \Delta_{\text{EI},j}, A_{\text{ME},j}),
\end{equation} 

\noindent where $s$ denotes the stage within each breathing period: $\text{EE}$ for the end of exhale (or beginning of inhale), $\text{EI}$ for the end of inhale (or beginning of exhale), and $\text{ME}$, $\text{MI}$ for the 2 intermediate points between EE and EI. For simplicity, we omit the redundant $\Delta_{\text{ME},j}$ and $\Delta_{\text{MI},j}$ time coordinates, since they are equal to $\Delta_{\text{EI},j}/2$ and $\Delta_{\text{EE},j}/2$, respectively.

\begin{figure}[]
\begin{subfigure}[t]{.5\textwidth}
  \centering
  \includegraphics[width=.8\textwidth]{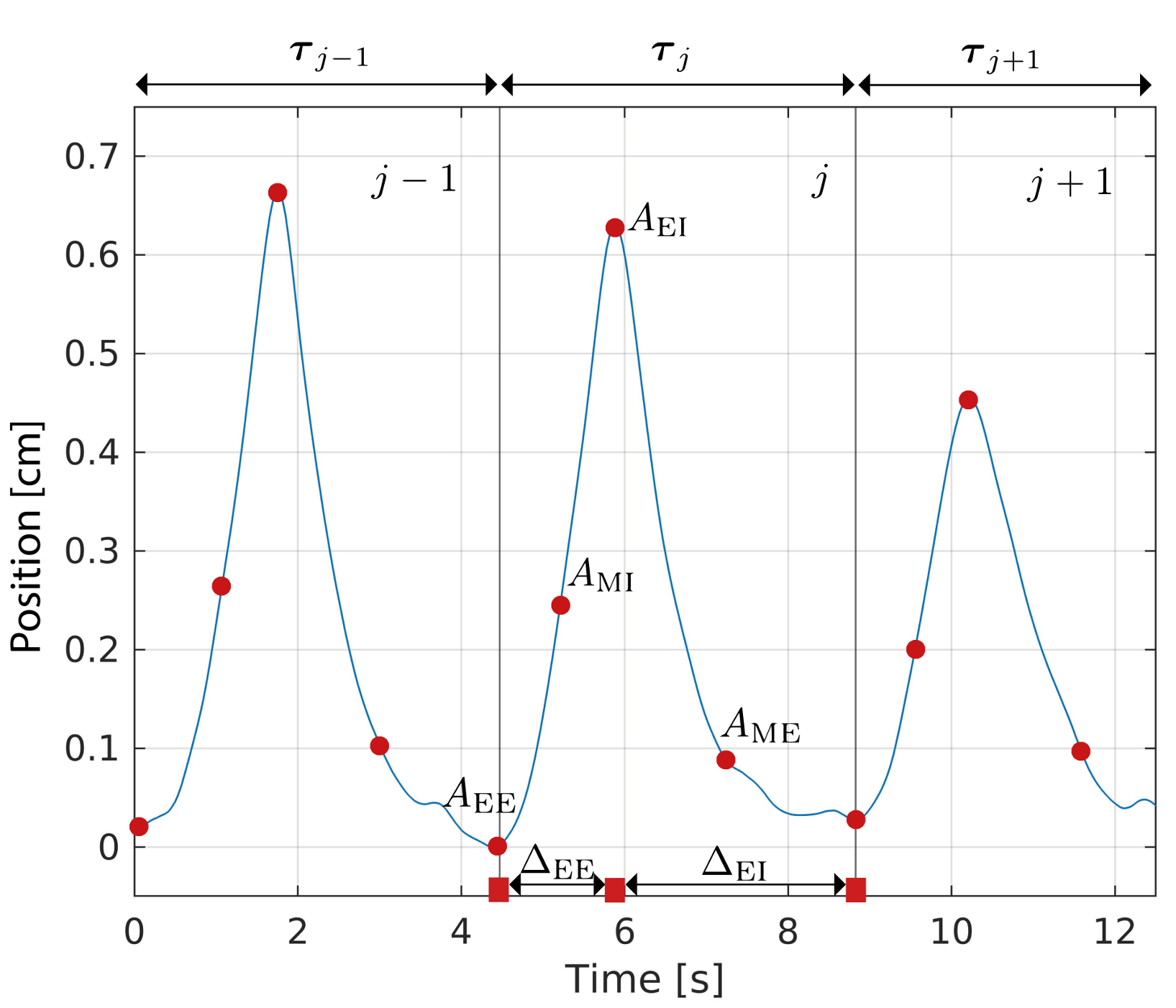}   
  \caption{Times series discretization.}
  \label{fig:param}
\end{subfigure} \qquad
\begin{subfigure}[t]{.5\textwidth}
  \centering 
  \includegraphics[width=.8\textwidth]{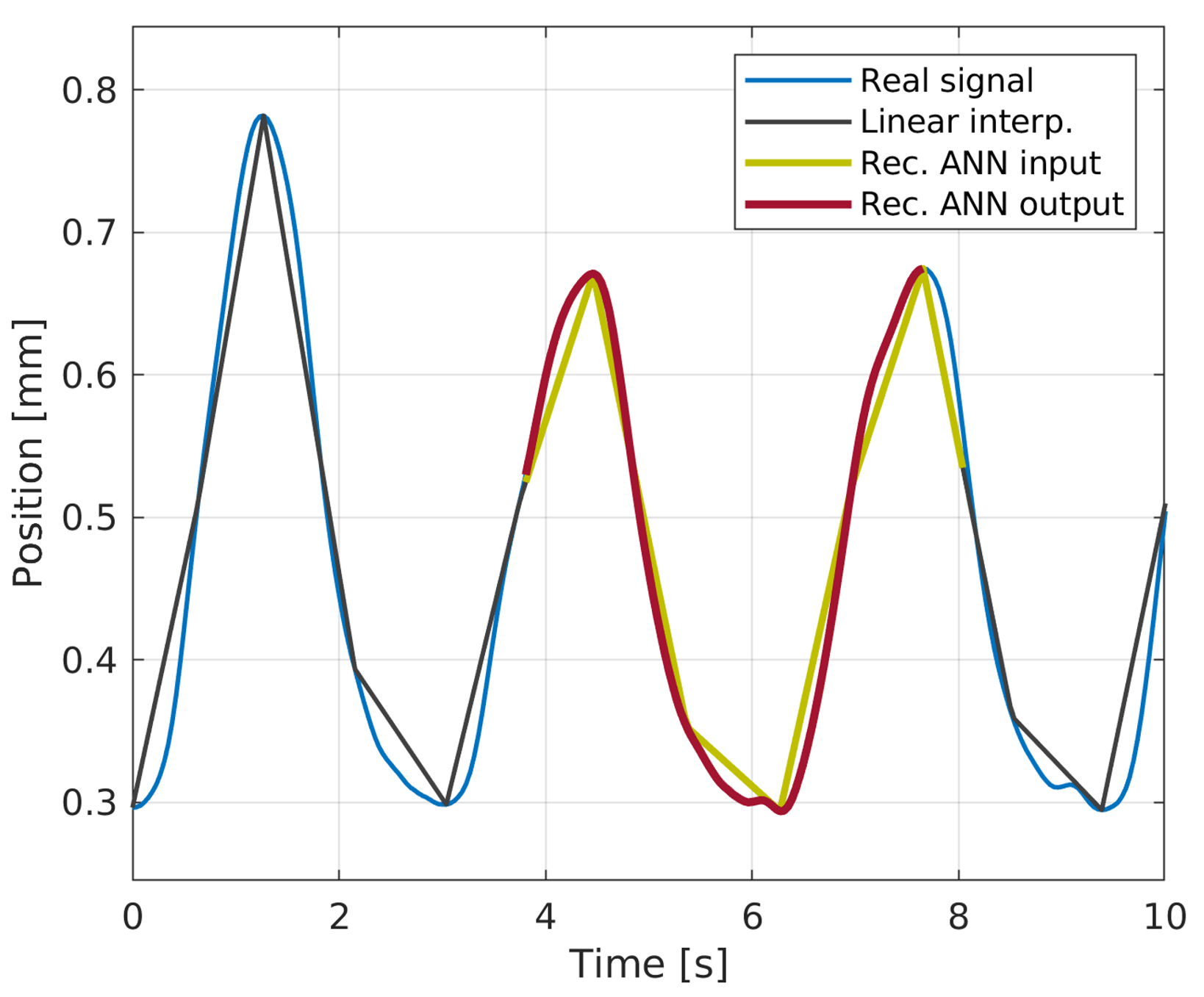}
  \caption{Time series reconstruction.}
  \label{fig:recann}
\end{subfigure}
\caption{ (a) Discretization of a breathing signal into periods and time-position points. In practice, the time series is discretized into a pair of time-position coordinates that are concatenated for a number of periods covering a certain desired time. (b) Transformation of the vector $\bm{x}$ into a time series. An additional ANN is trained to generate realistic breathing signals from linearly interpolated time series.}	
\label{fig:vaae}
\end{figure}
	
Figure \ref{fig:param} displays a fragment of the time series and its discretization into time-position points. A breathing sample is obtained by concatenating consecutive periods for the desired length of the signal. Each sample is assumed to be i.i.d. and is characterized by a vector $\bm{x}= (\bm{\tau}_1, \bm{\tau}_2,..., \bm{\tau}_{N_T})\in\mathbb{R}^{N_T\times 6}$ formed by $N_T$ discretized periods. We use vectors of length $N_T=25$ to model shorter signals of 1 to 2 minutes, and $N_T=100$ for longer signals of several minutes corresponding to the typical duration of radiotherapy treatments. This compression step allows reducing the dimensionality of the breathing time series two orders of magnitude. 

The pre-processing step results in 36430 and 4468 breathing fragments for the GUH and EMC datasets, respectively. Each data sample is assigned a baseline shift label according to the slope of the signal: if the slope of a sample is above a certain threshold value, the breathing sample is labeled as upwards baseline shift. Likewise, if the (negative) slope is below the threshold, the data point is labeled as downwards baseline shift. The threshold values correspond to the 7.5 upper and lower percentile of the distributions of slopes in the GUH dataset.

\subsection*{Convolutional filters.} We use one-dimensional convolutional layers for both the encoder and decoder models under the assumption that these provide the encoder and decoder with powerful feature extractors  that exploit the order in time and local structure of the periods. A one-dimensional discrete kernel convolution operation (denoted as $\bm{x}*\mathcal{K}$) over an input $\bm{x}\in\mathbb{R}^{N_T\times 6}$ with $N_T$ time-steps and 6 channels for the different time and position values, using a kernel $\mathcal{K}\in\mathbb{R}^{K\times 6}$, consists of sliding the kernel matrix through the different $j$ time-steps and computing

\begin{equation}
	(\bm{x}*\mathcal{K})(j)= \sum_{k=1}^K\sum_{h=1}^6 \mathcal{K}_{k,h}\:x_{j-k,h}.
\end{equation}

\subsection*{Patient-specific models.} To investigate the potential and limitations of signal modeling with probabilistic autoencoders, we first apply the standard VAE and AAE algorithms to model breathing signals from individual patients in the dataset separately. We train both the AAE and VAE frameworks using an isotropic Gaussian prior distribution $p(\bm{z})=\mathcal{N}(\bm{z};\bm{0},\bm{I})$. For the VAE, the parameter $\beta$ in \egyref{eq:elbo} is normalized with respect to the input dimension $M$ and latent dimension $N$ (which vary per model) as $\beta_n = (M/N)\beta$. 80\% of the patient data is used to train the model, while the remaining 20\% is equally split into a validation and a test set. Both the encoder and decoder consist of 4 convolutional layers and 2 fully-connected layers. Details about training and the architecture of the different models in the VAE and AAE are shown in Appendix \ref{app:aae}. After training the models, the input vector $\bm{x}$ can be reconstructed by sampling the inference model $q_{\phi}(\bm{z}|\bm{x})$ to obtain $\bm{z}$, and then sampling the decoder. Artificial breathing signals can be obtained by decoding random samples from the prior $p(\bm{z})$.

\subsection*{Evaluating patient-specific models.}A good model is capable of reconstructing unseen signals and generates artificial signals that distribute according to the training data. We perform several tests to asses the  generative performance of the patient-specific model:

\begin{itemize}
	\item Analyzing reconstruction error. To assess the reconstruction and generalization performance of the patient-specific models, we evaluate the reconstruction error of signals from the test set. For a fixed encoder and decoder architecture, we investigate the effect that varying the dimensionality of the latent space has on the reconstruction error of unseen test data. We verify and quantify the advantages of using convolutional layers by training models purely based on fully-connected layers and compare them to the one-dimensional convolutional models in terms of reconstruction performance.
	\item Assessing the generative performance. To determine if the model captures the data distribution, we train a classifier to distinguish between reconstructed and artificial samples from the model. Based on the same reasoning as in \cite{vqvae2}, we use reconstructed data instead of the original input vectors, since the compression through the latent space usually removes high-frequency noise in the original data that can be easily used by the classifier to distinguish samples. The classifier performance is evaluated for different latent space dimensionalities. 
	\item Investigating the structure of the latent space. The presence of "empty" regions in the latent space where no encodings $\bm{z}$ data are observed often results in low quality and variability of training samples. To determine the presence of empty regions, we evaluate the distribution of the distance between neighboring $\bm{z}$ from the dataset. Additionally, we evaluate possible mismatches between the aggregated posterior and prior distribution by comparing the distribution of the L2 norm of the encodings of the training samples and the samples from the prior.
\end{itemize}

\subsection*{Joint semi-supervised models of breathing irregularities.} We apply the semi-supervised SAAE framework to model and classify baseline shift breathing irregularities, which are gradual downward or upward shifts of the exhale position. First, we perform two simple experiments using an analytical dataset that contains simplified sinusoidal breathing signals. In the first experiment (S1), we vary only the slope of the signals. In the second experiment (S2), we also modify the period and amplitude. The goal of S1 and S2 is to determine whether it is possible to obtain good models that classify and generate signals with upward or downward shift, or no shift at all (regular signals). 

In the third experiment, we train the SAAE model using real breathing signals, and investigate the number of labeled samples needed to obtain accurate classification. All models are trained using the GUH dataset as the training set (with 10\% as validation data) and tested on the EMC dataset.

\subsection*{Evaluating breathing irregularities models models.} The evaluation of the joint models is based on the F1-score, which was first introduced by van Rijsbergen \cite{f1} and is computed as $\text{F1}(p,r)=2pr/(p+r)$,  where $p$ and $r$ are the per-class precision and recall, respectively. For a given class, the precision is the proportion of correctly predicted samples over the total number of examples labeled as such class, while the recall is the fraction of correctly predicted samples over the total number of true samples for the given class. For multi-label classification, the macro F1-score (mF1) can be used, which is the average of F1-scores for the different classes. The baseline shift breathing irregularity models are tested with regards to both their classification and generative performance.

\begin{itemize}
	\item Assessing classification performance. The discriminative performance (i.e., the ability to label signals having upward, downward or no baseline shift) is evaluated by comparing the classification accuracy of SAAE models to other neural network models purely optimized for classification. Specifically, convolutional neural network and fully-connected neural network discriminators are trained using a labeled subset of the training data. This additional convolutional classifier is similar to the encoder and inspired by state-of-the-art one-dimensional convolutional ECG models in \cite{c1} and \cite{c2}. We investigate how the number of labeled examples used during the supervised phase of training affects the classification accuracy of the SAAE by comparing its mF1-score to that of pure classifier networks. 
	\item Evaluating generative performance. Inspired by \cite{cas} and \cite{vqvae2}, we evaluate the generative performance by calculating the Classification Accuracy Score (CAS), which allows to gauge whether the model generates realistic and varied samples. The CAS is obtained by training a discriminative model on data generated by the model, and evaluating the mF1-score on the real data test set. 
	\item Analyzing the reconstruction error. Additionally, we evaluate the reconstruction performance of the model on GUH and EMC test data using 15 and 30 latent variables.
\end{itemize}

\subsection*{Time series reconstruction.} The output vectors $\hat{\bm{x}}$ from the models have the same structure as the discretized input vector. Therefore, they must be transformed back into a time series by reconstructing the position values between two consecutive points in $\hat{\bm{x}}$. A first order approximation is a simple linear interpolation between the four position points in each cycle, which requires little time but lacks accuracy.

Alternatively, we reconstruct a realistic breathing time series using an additional feed-forward neural network, which we denote \textit{reconstruction ANN}. The input is the linearly interpolated series, and the ANN learns a general mapping from the linear time series into realistic shapes. The input for the reconstruction ANN is no longer a vector of dimension $M=6\times N_T$, but a fragment of 120 position values (see \figref{recann}). The number 120 is a hyperparameter that is selected from a set of different candidate lengths. The output of the ANN is the first 100 transformed values of the input series. By adding 20 extra positions, the network achieves higher accuracy without discontinuities during concatenation of consecutive fragments. Further description of the ANN architecture is included in \ref{app:rec}.

The training data for the reconstruction ANN consists of slices with 120 elements of position values from the recorded breathing signals, and the corresponding linear interpolations. During training, the input and output slices are normalized to the interval [0,1]. A single general ANN would allow to reconstruct the time series from any patient in the population and make the process highly scalable. We investigate whether it is possible to train a general reconstruction ANN using only a subset of the data (either data from a single patient or a subset of data from all the patients). For this, we train the reconstruction ANN using (i) data from one patient (referred to as PatBR model from on) and (ii) a subset of data from the GUH data (referred to as PopBR model), while both models are tested using the EMC dataset. The PatBR is trained using a single patient from the GUH dataset, while the PopBR is trained on 10\% of the GUH dataset, instead of on all available samples. This is due to the fact that, unlike with the AAE, VAE and SAAE vector inputs, the training dataset for the reconstruction task consists of few million fragments of the breathing time series (vectors with 120 position values) obtained from linear interpolation of the generated vectors.
 
\section{Results}

\subsection*{Patient-specific models.} 
\begin{figure}[]
	\centering
	\includegraphics[width=\textwidth]{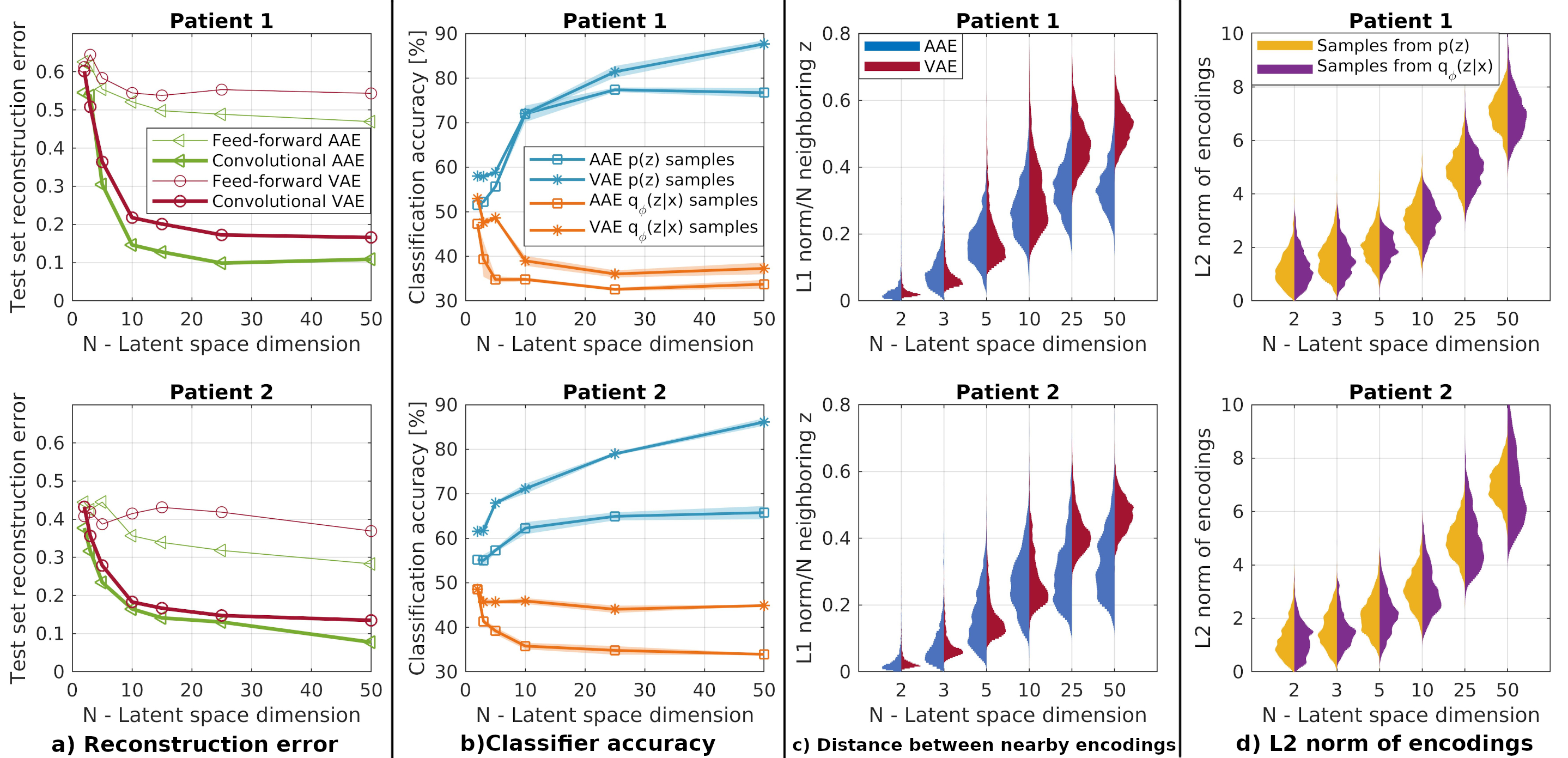} 
	\caption{Summary of the patient-specific model evaluation. (a) Reconstruction error on the test set for different latent space dimensionalities N. (b) Performance of an additional classifier trained to distinguish samples from the dataset from artificial samples from the model. Shaded regions represent the standard deviation around the mean (solid). (c) Distribution of the distance between neighboring encodings, for the AAE (blue) and VAE (red). The L1 norm distance is normalized by dividing by the latent space dimensionality. (d) Distribution of the L2 norm of the real data encodings $\bm{z}$ (yellow) and sampled encodings from the prior distribution (purple).	
	\label{fig:lsdim}}
\end{figure}

\begin{figure}[]
	\centering
	\includegraphics[width=\textwidth]{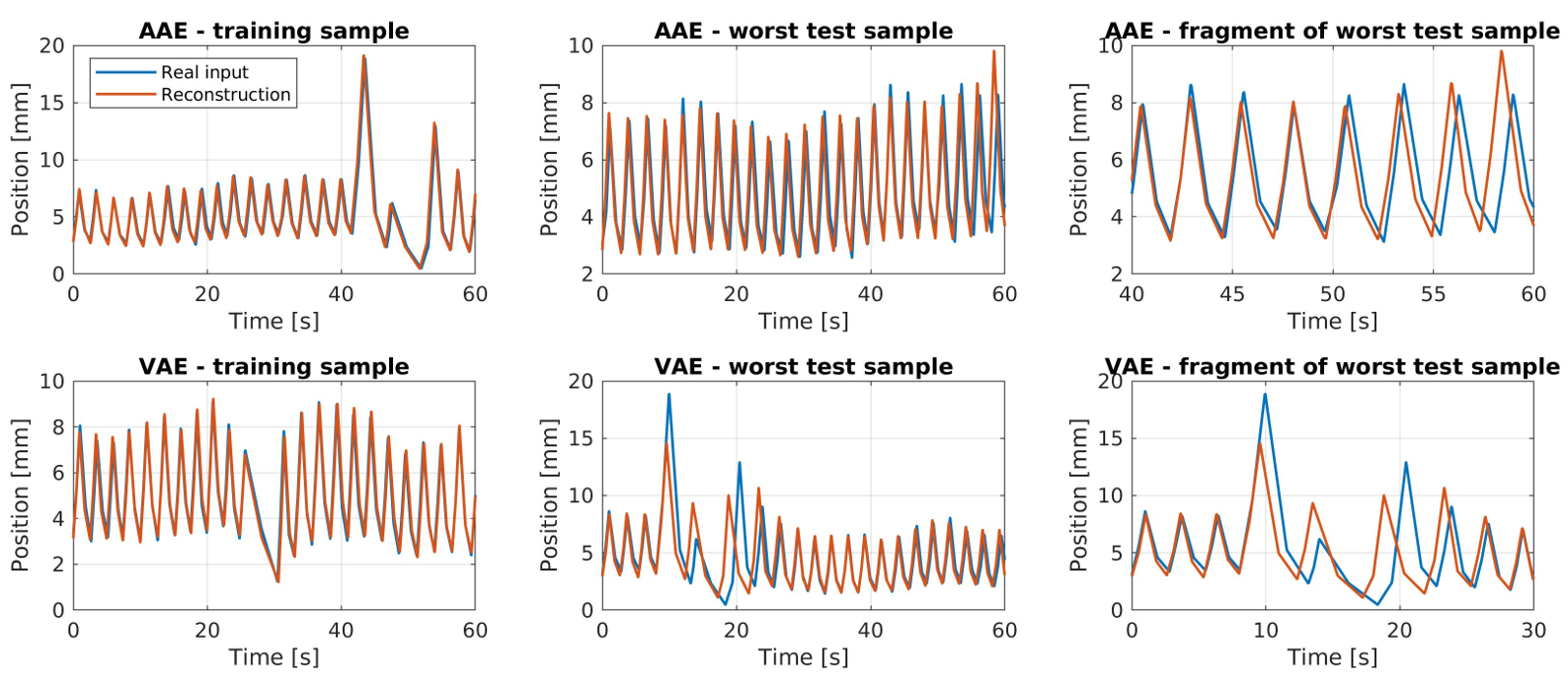} 
	\caption{(Top row) Reconstruction of breathing signals from AAE patient-specific models and (bottom row) VAE-based models for (left) a sample from the training set, (middle) the worst performing sample from the GUH test set, and (right) and a fragment of the worst reconstructed GUH test sample, with the highest reconstruction error. The discretized reconstructed signals are linearly interpolated and transformed back into a time series.}		
	\label{fig:rec}
\end{figure}

The results of the evaluation of the AAE and VAE patient specific models in terms of reconstruction and generative performance are shown in \figref{lsdim} for 2 randomly selected patients. The models for the first and second random patient were trained using 1890 and 2653 samples, respectively. \figref{lsdim}a displays the reconstruction error on unseen test set data for different latent space dimensionalities. The error values are re-scaled to the interval [0,1] to facilitate comparison, 1 corresponding to the maximum error achieved at weight initialization. We compare the error achieved by models based on one-dimensional convolutional architectures and models purely based on fully connected layers. Although the error always decreases with increasing latent dimension $N$, the convolutional architectures notably increase the accuracy in the reconstruction. For qualitative evaluation, \figref{rec} shows reconstructions of the original inputs using a convolutional model with a 5-dimensional latent space ($N=5$).

The generative performance is shown in \figref{lsdim}b, depicting the accuracy of a CNN classifier trained to distinguish reconstructed data points from artificial samples generated by models with varying latent dimensionality. We plot the average and standard deviation of 3 different classifiers trained on distinct artificial data. The data is generated either by sampling the prior $p(\bm{z})$, or by taking $\bm{z}$ encodings in the vicinity of $q_{\bm{\phi}}(\bm{z}|\bm{x})$, where the latter cover a much smaller domain of the latent space. The auxiliary classifier performs worse when distinguishing real and AAE samples, hinting that these better capture the distribution of the data. Note that the binary cross entropy loss values are almost always above 1 for the $p(\bm{z})$ classifier, which indicates the presence of uncertainty and significant miss classification errors.

To study the structure of the latent space \figref{lsdim}c shows the distribution of the distance between neighboring encodings. Since the $L^n$-norm distance metric always increases with the number of latent dimensions $N$, we divide the L1 norm between nearby $\bm{z}$ by the latent space dimensionality. The plotted distributions indicate that the AAE encodings are more evenly distributed. This, together with the fact that the classifier in \figref{lsdim}b struggles to distinguish real signals from samples in the vicinity of $q_{\bm{\phi}}(\bm{z}|\bm{x})$ hints that the latent space is more compact in the AAE-based models. On top of that, the AAE algorithm seems to be a more effective latent space regularizer, whose models have a latent space that closely resembles the prior distribution. This is deduced from \figref{lsdim}d, where the distribution of the L2 norm of the encodings is compared to the distribution of the L2 norm of samples from the prior. The results suggest a possible relationship between more compact and similar to the prior AAE latent space and the lower classifier performance for AAE samples in \figref{lsdim}b. Appendix \ref{app:ls} directly shows the distribution of the encodings, as well as a visualization of how data is organized in the low-dimensional latent space.

\subsection*{Semi-supervised baseline shift population models.} We first evaluate the effect of slope, period and amplitude variations on the classification accuracy by using an artificial dataset based on sinusoidal signals. The SAAE models achieve a mF1-score of 100\% in S1 by using as little as 300 labeled examples during the supervised classification phase. Adding period and amplitude variability to the sinusoidal signals in S2 results in additional difficulty, and the models need 1500 labeled examples (around 4\% of the training data points) in order to achieve null classification error.

\begin{table}[]
	\centering
	\caption{Overview of the classification, generation and reconstruction performance of the SAAE semi-supervised models, using 15 and 30 latent variables. The models use 4\% or 12\% of the training data during supervised classification phase, which corresponds to 1500 and 6000 data points, respectively. For the classification and generation results, we show the mean mF1-score and the standard deviation obtained from training 3 independent classifiers on different generated data. The relative reconstruction error is expressed as a percentage, where 100\% corresponds to the maximum error corresponding to a model with randomly initialized weights.}
	\begin{tabular}{@{}ccccc@{}}
		\toprule
		& \multicolumn{4}{c}{\textbf{Classification (classifier mF1-score [\%])}}                                             \\
		& \multicolumn{2}{c}{\textbf{GUH test set}}  & \multicolumn{2}{c}{\textbf{EMC test set}}  \\
		& \textbf{4\% labels} & \textbf{12\% labels} & \textbf{4\% labels} & \textbf{12\% labels} \\ \midrule
		\textbf{Feed-forward}        & 96.60 $\pm$ 2.09    & 98.64 $\pm$ 0.17     & 87.32 $\pm$ 2.33    & 93.77 $\pm$ 0.88     \\
		\textbf{CNN}        & 96.43 $\pm$ 1.46    & 98.64 $\pm$ 0.34     & 90.71 $\pm$ 1.20    & 93.92 $\pm$ 0.81     \\
		\textbf{SAAE, N=15} & 94.04 $\pm$ 2.25    & 99.00 $\pm$ 0.67     & 93.04 $\pm$ 0.59    & 96.28 $\pm$ 0.35     \\
		\textbf{SAAE, N=30} & 95.18 $\pm$ 1.42    & 98.14 $\pm$ 0.93     & 92.82 $\pm$ 1.65    & 94.39 $\pm$ 1.80     \\ \midrule
		& \multicolumn{4}{c}{\textbf{Generation (CAS mF1-score [\%])}}                                           \\
		& \multicolumn{2}{c}{\textbf{4\% labels}}    & \multicolumn{2}{c}{\textbf{12\% labels}}   \\ \midrule
		\textbf{SAAE, N=15} & \multicolumn{2}{c}{83.00 $\pm$ 0.49}       & \multicolumn{2}{c}{91.76 $\pm$ 0.68}       \\
		\textbf{SAAE, N=30} & \multicolumn{2}{c}{70.25 $\pm$ 0.47}       & \multicolumn{2}{c}{76.75 $\pm$ 1.68}       \\ \midrule
		& \multicolumn{4}{c}{\textbf{Reconstruction (relative error [\%])}}                                             \\
		& \multicolumn{2}{c}{\textbf{GUH test set}}  & \multicolumn{2}{c}{\textbf{EMC test set}}  \\
		& \textbf{4\% labels} & \textbf{12\% labels} & \textbf{4\% labels} & \textbf{12\% labels} \\ \midrule
		\textbf{SAAE, N=15} & 19.98 $\pm$ 0.28    & 18.90 $\pm$ 1.27     & 45.74 $\pm$ 5.44    & 40.56 $\pm$ 3.75     \\
		\textbf{SAAE, N=30} & 18.82 $\pm$ 0.96    & 18.50 $\pm$ 0.28     & 37.95 $\pm$ 3.10    & 36.21 $\pm$ 4.83     \\ \bottomrule
	\end{tabular}
	\label{tab:saae}
\end{table}

Based on these results, we train a baseline shift model using real data. The performance and added benefits of jointly classifying and modeling breathing signals are evaluated by assessing the classification accuracy, generation variability and the reconstruction error. The classification performance is assessed by comparing the SAAE models to purely discriminative models specifically trained to classify baseline shifts using a subset of the available labels. Specifically, a feed-forward and convolutional classifiers were trained using 4\% and 12\% of the GUH training labeled data. \tabref{saae} shows that our model with 15 latent variables outperforms both architectures by around a 2.5\%, achieving a mF1-score of $93.04\pm 0.59$ and $96.28\pm 0.35$ on the unseen test EMC dataset when trained with 4\% and 12\% of the labels, respectively. Taking into account the standard deviation, the classification accuracy on GUH test data is on par with that of the feed-forward and CNN classifiers. 

The generative performance and sample variability are evaluated with the CAS mF1-score. A CNN classifier is trained using 36430 randomly generated samples from the SAAE model, which allows a fair comparison with the model trained using the real GUH data. The classifier is then evaluated on EMC data, achieving a remarkable $91.76\pm 0.68$ mF1-score for the model with 15 latent variables trained with 12\% of the labels, which is not far from the performance of the feed-forward and CNN classifiers trained with real data observed in \tabref{saae} (Feed-forward and CNN rows). As with the patient-specific models, the generative performance significantly degrades for higher latent space dimensionality.

Finally, the reconstruction error on test set data is shown in \tabref{saae}. As with the patient-specific models, the error is expressed relative to the maximum error corresponding to predictions from a randomly initialized model. The models perform similarly regardless of the latent dimension. Higher latent space dimensionality seems to beneficial in the complicated task of reconstructing EMC samples that follow a different distribution, where the models achieve similar reconstruction performance to the feed-forward patient-specific models in \figref{lsdim}a.

\subsection*{Time series reconstruction.} 
\begin{figure}[]
	\centering
	\includegraphics[width=\textwidth]{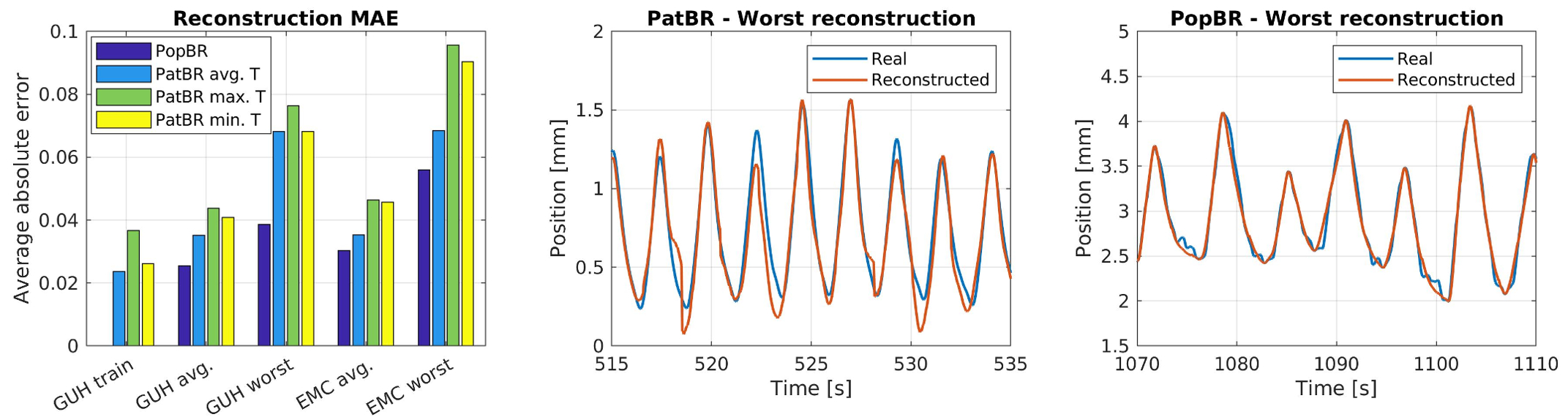} 		
	\caption{(Left) Average absolute error achieved by the PatBR and PopBR models in the reconstruction of breathing time series. The error is shown for the training patient(s), the worst-performing patient and the entire set of patients present in  each of the GUH and EMC datasets. (Middle) Reconstruction of the EMC signal fragment with highest error, using the PatBR model trained with data from the patient with maximum amplitude. (Right) Worst-performing reconstruction over all the EMC dataset using the PopBR model.  }
	\label{fig:brrec}
\end{figure}

Three different PatBR models are trained using the data from three patients: the patients with the largest and lowest breathing period in the dataset, and one of the patients with an average period. From these PatBR models, the former (largest period) achieves the largest error, precisely on signals of the patient with the lowest period. For each of the PatBR models and the PopBR model, a comparison of the average absolute error (average L1-norm) on the training set, the test set and the worst performing patient from the test set is shown in the left plot of \figref{brrec}. The average absolute error is calculated as the average L1-norm $\abs{\bm{w}_{\text{real}}-\bm{w}_{\text{rec}}}$  between all position points in the recorded and reconstructed time series vectors $\bm{w}$. The middle and right plots in \figref{brrec} show the worst EMC test sample reconstruction from the PatBR and PopBR models, respectively.

\section{Discussion}

\subsection*{Reconstruction accuracy and effect of convolutions.} The standard VAE, standard AAE and  SAAE architectures result in breathing models that capture the variability of respiration through few latent variables, as opposed to approaches that use implicit adversarial models \cite{gen3} \cite{gen4}. The models are easy to sample and the decoders generate realistic breathing samples. The convolutional layers result in 25\% reduction of the reconstruction error on test data. AAEs outperform standard VAE models in reconstruction, generalization and generative performance. Much of the AAE success seems to be related to their more compact latent space: their aggregated posterior distributions are closer to the prior, and their encodings are more evenly spaced, as seen in \figref{lsdim}c and \figref{lsdim}d. The problem of aggregated posterior-prior mismatch in VAEs is not new, and our findings support previous studies \cite{hp} \cite{pm1} \cite{pm2}.

\subsection*{Effect of latent space dimensionality.} For the set of all possible models, the reconstruction performance is in theory independent of the latent dimension. Very powerful autoencoders with deep encoders and decoders could perfectly reconstruct the input using as few as one latent dimension, but this is not observed in practice. In general, the performance can be practically improved by adding more latent variables or increasing the capacity of the model. However, it has been observed that very powerful decoder architectures tend to ignore the information encoded in $\bm{z}$ \cite{pc}\cite{infovae}\cite{vla}. In concordance with \figref{lsdim}, adding dimensions helps, especially in low-dimensional latent spaces. Nevertheless, there is a certain latent space dimensionality beyond which adding more latent units seems to add little information. For the VAE, this may manifest as "inactive latent variables", where some latent units remain equal to the prior distribution during the whole training process \cite{iwae}\cite{laddervae}. For the specific case of breathing and given the presented encoder and decoder convolutional architectures, the limit seems to be around 10 latent variables. This is supported by the fact that the test reconstruction error and classifier performance plateau around $N=10$ in \figref{lsdim}.

\subsection*{Semi-supervised models.} Even though SAAE models are mainly trained to reconstruct breathing signals, they outperform pure discriminative architectures based on state-of-the-art one-dimensional convolutional models \cite{c1}\cite{c2}. The fact that a single model can (better) classify and selectively sample types of signals is a novelty with respect to previous architectures that specialize in only one of such tasks \cite{gen1} \cite{gen4}. One interesting remark is the fact that the SAAE models with 15 latent variables outperform the SAAE models with 30 latent variables in the classification task. In general, increasing the number of latent variables means that less information about the input is encoded per latent variable. We hypothesize that  some of the information encoded in $\bm{y}$ may leak into the style variables $\bm{z}$ and cause loss of accuracy for increasing latent space dimensions. However, this should be confirmed in future research.

The generative performance of the SAAE models degrades with increasing latent dimensionality. As in the patient-specific models, we hypothesize that this is the result of an "emptier" latent space with larger distance between encodings. Additionally, SAAE models perform similarly to the patient-specific models in terms of reconstruction and generalization on test samples from the same distribution, as indicated by the reconstruction error on GUH test samples. Although the reconstruction accuracy significantly decreases, the SAAE models also perform reasonably well in the much more complicated task of generalizing to test samples from the EMC dataset with different distribution, and their reconstruction error is on par with feed-forward patient-specific models (\figref{lsdim}a).

\subsection*{Time series reconstruction accuracy.} The PopBR reconstruction ANN consistently outperforms the single patient PatBR networks and opens the door to using a single model to reconstruct breathing signals for any patient. PatBR models fail to reconstruct time series from other patients, especially when they are evaluated on patients whose period significantly differs from that of the samples used for training, as seen in the left plot of \figref{brrec}. The generalization error of the PopBR model is very low and it provides accurate reconstructions for patients whose breathing signal was recorded in a different location and machine. The error could in principle be further decreased by training a specific PatBr for each specific patient, at the expense of slightly longer computation time.

\subsection*{Usefulness of breathing models.} The models presented in this paper can be applied to a wide range of tasks involving generation and classification. First, the AAE and VAE can be used to capture the variability in breathing of a patient and generate artificial patient samples. These can be incorporated into treatment design and delivery in order to obtain robust treatments beforehand, e.g., obtaining proton therapy treatments robust against breathing movements that result in desired clinical outcomes, or estimating the likelihood that a patient will present a certain type of breathing. 

The SAAE framework can in principle be applied for computer aided diagnosis of breathing abnormalities, as well as for dataset augmentation when the available data for a patient is scarce. An example is classifying breathing irregularities and generating additional samples that present the identified irregularity. One of the advantages of training the proposed framework in a semi-supervised way is the possibility to build such models requiring only a small subset of labeled data.

Our models can in principle be applied to any other kind of biomedical data that shows a repetitive or periodic structure, much like a breathing signal is composed of well-defined randomly varying periods with changing amplitude. To our knowledge, some of these signals could be ECG, electroglottograph (EGG), magnetoencephalography (MEG) or magnetocardiography (MCG). The added advantage of our generative approach with respect to other models in the literature that do not explicitly model the data distribution such as \cite{gen1} or \cite{gen2} is the possibility to map the data samples to specific regions or classes in latent space enabling classification and generation of data by sampling $\bm{z}$ from the desired regions. 

\subsection*{Limitations.} A notable drawback is the uninformative prior $p(\bm{y})$ in the semi-supervised model, which assumes no previous knowledge about the proportion between different classes. For cases when there is class imbalance, i.e., many more samples of regular breathing compared to irregular breathing, using such uninformative prior may result in the model miss-classifying some samples in order to match the uniform prior. The solution to this problem is dataset-dependent approach and involves determining the naturally occurring proportion of classes. 

\subsection*{Computational cost.} An important advantage of the presented methodology is the fact that it achieves feasible compute times. We reduce training times by using Graphics Processing Units (GPUs), which are needed to train the presented convolutional architectures due to the requirements of the latest version of the Tensorflow package \cite{tf}. We perform most of the training using an NVIDIA\textsuperscript{\tiny\textregistered} Tesla\textsuperscript{\tiny\textregistered} K80, and the training times vary around 10 minutes for the VAE and AAE patient-specfic models, 30 minutes for the reconstruction PopBR and PatBR models, and 20 minutes for the SAAE models. Generating and classifying breathing samples is almost instantaneous.

\section{Conclusion} 
We present a semi-supervised algorithm based on the AAE that allows simultaneous classification and generation of biomedical signals within a single framework, using few labeled data points. The resulting models classify signals with greater accuracy than discriminative models specifically trained for classification; are easy to sample, and compress the data into a reduced latent space with few independent parameters with known probability distributions. We show that 10 of such latent variables are able to capture most of the variation in the data and achieve excellent reconstruction and generation of samples. For the particular case of breathing, we demonstrate that the adversarial objective used in AAEs is a better regularizer of the latent space and overcomes some of the previously studied problems of the VAE framework.

Given the length of the input time series, we train the models on compressed input vectors containing information about the period and amplitude of the biomedical signal. The compressed output vectors of the generative models can be transformed back into a time series with the help of an additional reconstruction network. We demonstrate that a reconstruction model trained with the data of a single patient (PatBR) does not achieve good generalization when evaluated on other patients, and it is outperformed by a population model (PopBR) trained with a subset of the data of a population of patients. The population model is trained only once and can achieves great accuracy when applied to new unseen data. Even though we base our study on mechanical breathing signals, the framework shows potential applicability to simulation and diagnostic purposes using any other biomedical signal with a quasi-periodic structure.

\section{Acknowledgements}
This work is supported by KWF Kanker Bestrijding [grant number 11711], and part of the KWF research project PAREL. The authors are also thankful to Dr. Dennis Schaart for the interesting discussions regarding the manuscript, as well as Dr. Mischa Hoogeman and Dr. Steven Habraken for the insightful discussions and providing one of the datasets.

\subsection*{CRediT authorship contribution statement.}
\textbf{Oscar Pastor-Serrano}: Conceptualization, Data Curation, Methodology, Investigation, Formal Analysis, Visualization, Validation, Software, Writing - original draft. 
\textbf{Danny Lathouwers}: Supervision, Writing - review \& editing.
\textbf{Zoltán Perkó}: Conceptualization, Supervision, Methodology, Formal Analysis, Funding acquisition, Project administration, Writing - review \& editing.

\section{Code availability}
The code implementing training and evaluation of the AAE, VAE and SAAE, as well as the PopBR and PatBR reconstruction networks, is available at: \url{code_repo_url}.

\newpage
\appendix
\numberwithin{equation}{section}
\numberwithin{figure}{section}

\section{Evidence Lower Bound}
\subsection*{Deriving the ELBO.}\label{app:elbo} Even though there are different ways to obtain the ELBO, the most common derivation is based on Jensen's inequality. For a concave function such as the natural logarithm the Jensen inequality states that $$\log\:\big(\mathbb{E}[\bm{x}]\big)\geq\mathbb{E}\:[\log(\bm{x})].$$ Starting from the marginal likelihood of the probabilistic model, the expression of the ELBO can be obtained as

\begin{align}
\begin{split}
\label{eq:1}
	\log\:(p_{\bm{\theta}}(\bm{x})) ={}& \log\: \int_{\mathcal{Z}} p_{\bm{\theta}}(\bm{x},\bm{z})d\bm{z}
\end{split}\\
\begin{split}
\label{eq:2}
  ={}& \log\: \int_{\mathcal{Z}} p_{\bm{\theta}}(\bm{x},\bm{z}) \frac{q_{\bm{\phi}}(\bm{z}|\bm{x})}{q_{\bm{\phi}}(\bm{z}|\bm{x})}d\bm{z}
\end{split}\\
\begin{split}
\label{eq:3}
  ={}& \log\: \mathbb{E}_{\mathrm{\mathbf{z}}\sim q_{\bm{\phi}}(\bm{z}|\bm{x})} \Big[\frac{p_{\bm{\theta}}(\bm{x},\bm{z})}{q_{\bm{\phi}}(\bm{z}|\bm{x})}\Big]
\end{split}\\
\begin{split}
\label{eq:4}
  \geq{}& \mathbb{E}_{\mathrm{\mathbf{z}}\sim q_{\bm{\phi}}(\bm{z}|\bm{x})} \Big[\log\:\Big(\frac{p_{\bm{\theta}}(\bm{x},\bm{z})}{q_{\bm{\phi}}(\bm{z}|\bm{x})}\Big)\Big]
\end{split}\\
\begin{split}
\label{eq:5}
  ={}& \mathbb{E}_{\mathrm{\mathbf{z}}\sim q_{\bm{\phi}}(\bm{z}|\bm{x})} \Big[\log\:\Big(\frac{p_{\bm{\theta}}(\bm{x}|\bm{z})\: p(\bm{z}))}{q_{\bm{\phi}}(\bm{z}|\bm{x})}\Big)\Big]
\end{split}\\
\begin{split}
\label{eq:6}
  ={}& \mathbb{E}_{\mathrm{\mathbf{z}}\sim q_{\bm{\phi}}(\bm{z}|\bm{x})} [\log\:p_{\bm{\theta}}(\bm{x}|\bm{z})] - D_{KL}(q_{\bm{\phi}}(\bm{z}|\bm{x})||p(\bm{z})),
\end{split}
\end{align}

\noindent where the KL-divergence $D_{KL}$ is defined as 

\begin{equation}
	D_{KL}(p(x)||q(x)) = \int\log\Big(\frac{p(x)}{q(x)}\Big)\:p(x)\:dx = \mathbb{E}_{\text{x}\sim p(x)}\log\Big(\frac{p(x)}{q(x)}\Big).
	\label{eq:kl}
\end{equation}

\subsection*{Dissecting the ELBO.}
\label{app:mse}
The output of the probabilistic decoder is the likelihood conditional distribution $p_{\bm{\theta}}(\bm{x}|\bm{z})$. This distribution is represented as a multivariate Gaussian probability distribution with identity covariance matrix $p_{\bm{\theta}}(\bm{x}|\bm{z}) = \mathcal{N}(\bm{x};f_{\bm{\theta}}(\bm{z}),\bm{I})$, where the function $f_{\bm{\theta}}(z):\mathcal{Z}\rightarrow\mathbb{R}^M$ is parametrized with an ANN and represents the mean. The log-likelihood is formulated as

\begin{equation}
\log(p_{\bm{\theta}}(\bm{x}|\bm{z})) = \log\Bigg( \frac{1}{\sqrt{(2\pi)^M|\bm{I}|}} \exp \Big( -\frac{1}{2}(\bm{x}-f_{\bm{\theta}}(\bm{z}))^T\bm{I}^{-1}(\bm{x}-f_{\bm{\theta}}(\bm{z})) \Big)\Bigg) = C - \frac{1}{2}\|\bm{x}-f_{\bm{\theta}}(\bm{z})\|_2^2,
\end{equation}

\noindent where $C$ is a constant. The result has the same form as the squared error (SE), which is computed for the model output $\hat{\bm{x}}$ approximating the true output $\bm{x}$ as

\begin{equation}
\label{eq:mse}
\text{SE} = \|\bm{x}-\hat{\bm{x}}\|_2^2.
\end{equation}

Thus, minimizing the log-likelihood with respect to the parameters $\bm{\theta}$ (which is done by approximating the expectation $\mathbb{E}_{\mathrm{\mathbf{z}}\sim q_{\bm{\phi}}(\bm{z}|\bm{x})} \log(p_{\bm{\theta}}(\bm{x}|\bm{z}))$ by taking Monte Carlo samples for $\mathrm{\mathbf{z}}\sim q_{\bm{\phi}}(\bm{z}|\bm{x})$) yields the same result as minimizing the SE. On the other hand, when $p$ and $q$ are both Gaussian distributions, the KL-divergence can be computed in closed form. In our case the prior is $p(\bm{z})=\mathcal{N}(\bm{z};\bm{0},\bm{I})$ and the encoder distribution is $q_{\bm{\phi}}(\bm{z}|\bm{x}) = \mathcal{N}(\bm{z};\bm{\mu}(\bm{x}), \text{diag}\:\bm{\sigma(\bm{x})^2})$. For an N-dimensional latent space, the KL-divergence can be analytically computed as:

\begin{equation}
D_{KL}(q_{\bm{\phi}}(\bm{z}|\bm{x})||p(\bm{z}))=\frac{1}{2}\Bigg(-\sum_i^N(\log\sigma(\bm{x})_i^2+1)+\sum_i^N\sigma(\bm{x})_i^2+\sum_i^N\mu(\bm{x})_i^2 \Bigg).
\end{equation}

Note that the contribution of the KL-divergence to the ELBO scales linearly with the latent dimensionality, so an increase in the ELBO caused by an increase of the latent space dimensionality could in theory be compensated by increasing the variance of the approximated posterior $q_{\bm{\phi}}(\bm{z}|{\bm{x}})$ (lower KL-divergence per latent dimension).

\section{Adversarial variational objective}
\label{app:aaeloss}

AAEs do not exactly optimize the ELBO. This section describes the approximated variational objective in AAEs. In \cite{aae}, the authors propose to regularize the latent space by introducing a discriminator model, modeled also with an ANN with mapping function $d_{\bm{\xi}}(\bm{z}):\mathcal{Z}\rightarrow\mathbb{R}$ that outputs a single scalar logit. The discriminator is assumed to be capable of approximating any function. Given the encoder mapping  $g_{\bm{\phi}}(\bm{z}|\bm{x},\eta):\mathcal{X}\times H\rightarrow\mathcal{Z}$, and the approximated posterior distribution $q_{\bm{\phi}}(\bm{z}|\bm{x}) = \int_{H} \delta(\bm{z}-g_{\bm{\phi}}(\bm{x},\eta)) p(\eta) d\eta $, the adversarial regularization objective maximization can be formulated as

\begin{align}
	\begin{split}
		 \underset{\bm{\xi}}{\max}\:\:\mathbb{E}_{\mathrm{\mathbf{z}}\sim p(\bm{z})}[\log (S(d_{\bm{\xi}}(\bm{z})))]+\mathbb{E}_{\mathrm{\mathbf{x}}\sim \hat{p}_{data}(\bm{x})}\mathbb{E}_{\mathrm{\mathbf{z}}\sim q_{\bm{\phi}}(\bm{z}|\bm{x})}[\log (1-S(d_{\bm{\xi}}(\bm{z})))]
	\end{split}\\
	\begin{split}
		={}& \underset{\bm{\xi}}{\max}\:\:\int p(\bm{z})\log (S(d_{\bm{\xi}}(\bm{z})))d\bm{z}+\int\int \hat{p}_{data}(\bm{x})q_{\bm{\phi}}(\bm{z}|\bm{x})\log (1-S(d_{\bm{\xi}}(\bm{z})))d\bm{z}d\bm{x}
	\end{split}\\
	\begin{split}
		={}& \underset{\bm{\xi}}{\max}\:\:\int \Big[ p(\bm{z})\log 	(S(d_{\bm{\xi}}(\bm{z})))+\int \hat{p}_{data}(\bm{x})q_{\bm{\phi}}(\bm{z}|\bm{x})\log (1-S(d_{\bm{\xi}}(\bm{z})))d\bm{x}\Big]d\bm{z}.
	\end{split}
\end{align}

In the last step, we applied Fubini's theorem to change the order in the integration.  As in \cite{gan} and \cite{avb}, it can be shown that the discriminator achieves its optimum value at

\begin{equation}
	d_{\bm{\xi}}^*(\bm{z}) = \log(p(\bm{z}))-\log\Big(\int_{\mathcal{X}} q_{\bm{\phi}}(\bm{z}|{\bm{x}}) \hat{p}_{data}(\bm{x}) d\bm{x}\Big) = \log(p(\bm{z}))-\log(q_{\bm{\phi}}(\bm{z})).
	\label{eq:optdis}
\end{equation}

This follows from the fact that for any $(a,b)\in\mathbb{R}^2\setminus[0,0]$, a function that has the form $f(h) = a\log h + b\log(1-h)$ attains it maximum in $[0,1]$ at $h = a/(a+b)$. Thus, 

\begin{equation}
	S(d_{\bm{\xi}}^*(\bm{z})) = \frac{p(\bm{z})}{p(\bm{z})+\int_{\mathcal{X}} q_{\bm{\phi}}(\bm{z}|{\bm{x}}) \hat{p}_{data}(\bm{x}) d\bm{x}},
\end{equation}

\noindent which is equivalent to \egyref{optdis}. The ELBO in \egyref{elbo} can be reformulated based on the definition of the KL divergence in \egyref{kl} as

\begin{align}
	\begin{split}
		\label{eq:av0}
		\mathbb{E}_{\mathrm{\mathbf{x}}\sim \hat{p}_{data}(\bm{x})}[\log(p_{\bm{\theta}}(\bm{x}))]\geq{}&  	 \mathbb{E}_{\mathrm{\mathbf{x}}\sim \hat{p}_{data}(\bm{x})}\mathbb{E}_{\mathrm{\mathbf{z}}\sim q_{\bm{\phi}}(\bm{z}|\bm{x})} [\log(p_{\bm{\theta}}(\bm{x}|\bm{z}))]- \mathbb{E}_{\mathrm{\mathbf{x}}\sim \hat{p}_{data}(\bm{x})}[D_{KL}(q_{\bm{\phi}}(\bm{z}|\bm{x})||p(\bm{z}))]
	\end{split}\\
	\begin{split}
		\nonumber
		={}& \mathbb{E}_{\mathrm{\mathbf{x}}\sim \hat{p}_{data}(\bm{x})}\mathbb{E}_{\mathrm{\mathbf{z}}\sim q_{\bm{\phi}}(\bm{z}|\bm{x})} [\log(p_{\bm{\theta}}(\bm{x}|\bm{z}))]+
	\end{split}\\
	\begin{split}
		\label{eq:av1}
		& \mathbb{E}_{\mathrm{\mathbf{x}}\sim \hat{p}_{data}(\bm{x})}\mathbb{E}_{\mathrm{\mathbf{z}}\sim q_{\bm{\phi}}(\bm{z}|\bm{x})} [\log( p(\bm{z}))-\log(q_{\bm{\phi}}(\bm{z}|\bm{x}))].
	\end{split}
\end{align}

As described in \cite{aae}, the AAE algorithm replaces the last term in \egyref{av1} (regularization term, equivalent to the KL term) with "an adversarial procedure that encourages $q_{\bm{\phi}}(\bm{z}))$ to match to the whole distribution of $p(\bm{z})$". Mathematically, this translates into replacing the KL term with  $\mathbb{E}_{\mathrm{\mathbf{x}}\sim \hat{p}_{data}(\bm{x})}\mathbb{E}_{\mathrm{\mathbf{z}}\sim q_{\bm{\phi}}(\bm{z}|\bm{x})}[d_{\bm{\xi}}^*(\bm{z})]$, effectively approximating the variational bound as

\begin{align}
	\begin{split}
		\mathbb{E}_{\mathrm{\mathbf{x}}\sim \hat{p}_{data}(\bm{x})}\log\:(p_{\bm{\theta}}(\bm{x}))\geq{}&  	 \mathbb{E}_{\mathrm{\mathbf{x}}\sim \hat{p}_{data}(\bm{x})}\mathbb{E}_{\mathrm{\mathbf{z}}\sim q_{\bm{\phi}}(\bm{z}|\bm{x})} [\log(p_{\bm{\theta}}(\bm{x}|\bm{z}))]+\mathbb{E}_{\mathrm{\mathbf{x}}\sim \hat{p}_{data}(\bm{x})}\mathbb{E}_{\mathrm{\mathbf{z}}\sim q_{\bm{\phi}}(\bm{z}|\bm{x})} [d_{\bm{\xi}}^*(\bm{z})]
	\end{split}\\
	\begin{split}
		={}& \mathbb{E}_{\mathrm{\mathbf{x}}\sim \hat{p}_{data}(\bm{x})}\mathbb{E}_{\mathrm{\mathbf{z}}\sim q_{\bm{\phi}}(\bm{z}|\bm{x})} [\log(p_{\bm{\theta}}(\bm{x}|\bm{z}))]-D_{KL}(q_{\bm{\phi}}(\bm{z})||p(\bm{z})),
	\end{split}
\end{align}

\noindent where, compared to the ELBO in \egyref{av0}, the term $\mathbb{E}_{\mathrm{\mathbf{x}}\sim \hat{p}_{data}(\bm{x})}[D_{KL}(q_{\bm{\phi}}(\bm{z}|\bm{x})||p(\bm{z}))]$ is approximated with $D_{KL}(q_{\bm{\phi}}(\bm{z})||p(\bm{z}))$. As a result, the AAE translates into a modified variational objective that does not preserve the original formulation.

\section{Implementation details}

\subsection*{VAE architecture.}
\label{app:vae}
The architecture of the VAE models is shown in \figref{archs2}. We find that using BatchNormalization \cite{batchnorm} and Dropout \cite{dropout} between layers significantly improves convergence and results in significantly better generalization. The encoder contains one-dimensional max. pooling layers and the decoder uses dilation rates bigger than 1, which seem to positively affect reconstruction performance. A $\beta_n$ of 0.02 yields optimum balance between a Gaussian latent space that is closer to the prior and good reconstruction performance, with lower values slightly favoring more accurate reconstructions but aggregated posterior distributions with larger standard deviations that do not match the prior. We use a batch size of 256 samples and the Adam optimizer for training \cite{adam}, with learning rate $10^{-4}$.

\subsection*{Standard AAE architecture.}
\label{app:aae}
The architecture of the different models composing the AAE is shown in \figref{archs2}. For this framework, the order of the Batch Normalization and activation layers greatly affects convergence and stability during training, with Batch Normalization placed in between the activation and Dropout yielding the best results. Using Leaky ReLU activation functions with slope 0.1 in the discriminator also seems to help to stabilize training, in concordance with \cite{lrelu}. The models are trained using a batch size of 256 samples and the Adam optimizer with unequal learning rates: $2\cdot10^{-4}$ in the reconstruction phase and $10^{-4}$ for the discriminator. The squared error reconstruction loss is approximately 4 times lower than the cross-entropy loss used for the discriminator, and therefore multiplied by 4 during training.

\subsection*{Semi-supervised AAE architecture.}
\label{app:ssaae}
\figref{archs2} shows the architecture of the encoder, decoder and discriminator models for the semi-supervised modified AAE architecture. We find that Batch Normalization between layers in the encoder and decoder significantly boosts performance and helps stabilize training, as well as using unequal learning rates for the Adam optimizer: $10^{-4}$ in the reconstruction and supervised classification phase and $2\cdot10^{-4}$ for the discriminator. The models are trained using a batch size of 256 samples. As with the standard AAE architecture, the cross-entropy loss is approximately 4 times higher than the reconstruction error, and so the latter is equalized during training. We find that $\alpha$ values of around 5-10 significantly enhance classification when the number of labels is limited, while higher values do not improve and even hinder performance.

\subsection*{Reconstruction network.} 
\label{app:rec}
\begin{figure}[]
\centering
\includegraphics[width=\textwidth]{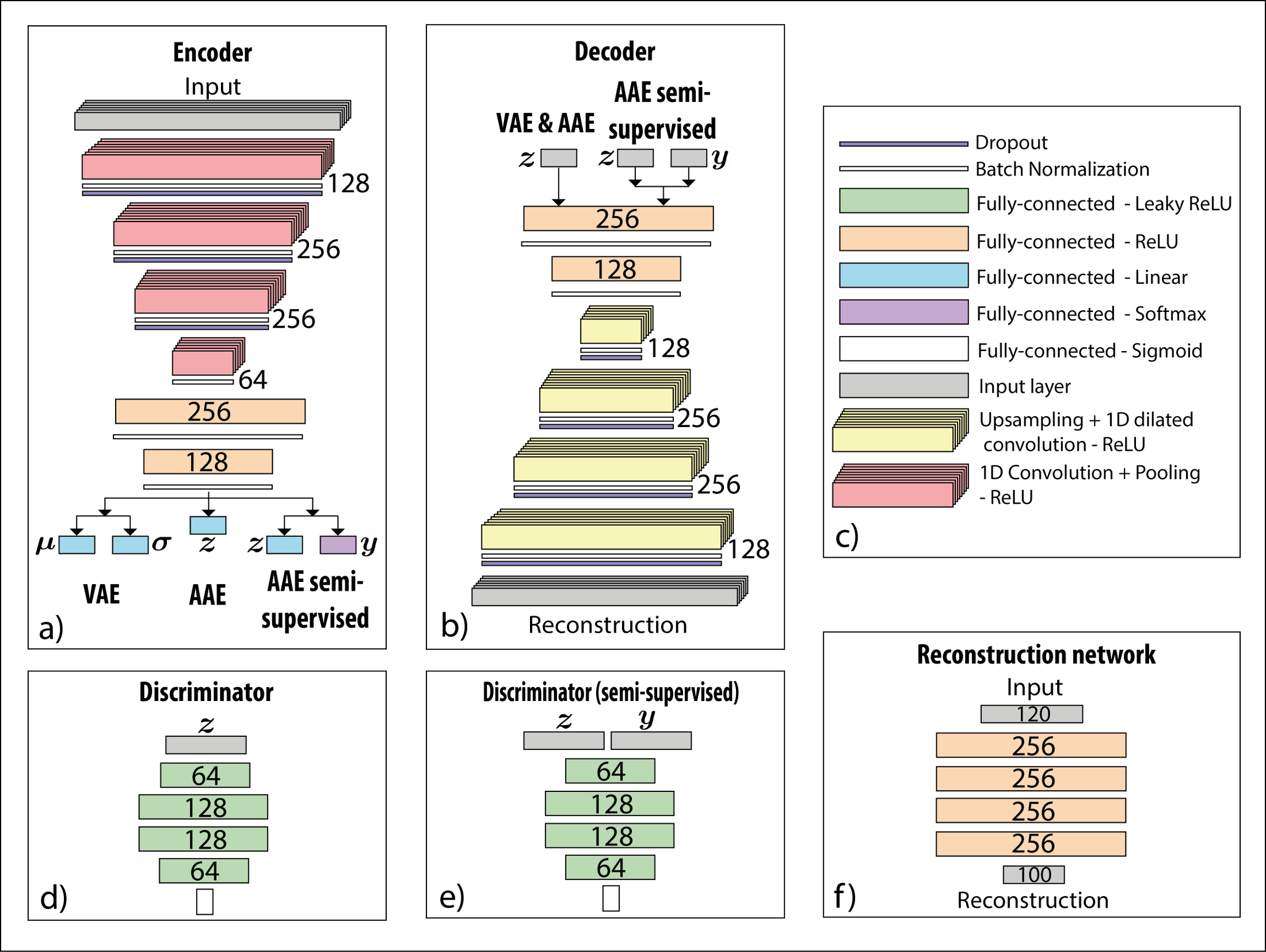} 		
\caption{Architecture of the different networks used in the AAE and VAE algorithms. (a) Convolutional encoder architecture with 4 one-dimensional convolutional layers and 2 fully-connected layers. A 1-D max-pooling layer follows each convolution, and Batch Normalization and Dropout with probability 0.1 are applied after each pooling layer. (b) Convolutional decoder architecture, with 2 fully-connected layers followed by 4 up-sampling dilated one-dimensional convolutional layers. Batch Normalization and Dropout with probability 0.3 follow each of the convolutions. (c) Color code for the layers used in the different models. (d) Discriminator architecture for the AAE, containing 4 fully-connected hidden layers followed by a sigmoid unit. (e) Discriminator for the SAAE. (f) Reconstruction network transforming the interpolated time series into realistic shapes.}
\label{fig:archs2}
\end{figure}

The architectures for the PatBr and PopBR models are identical and are shown in \figref{archs2}. The learning rate is set to $10^{-4}$ with a decay rate of $10^{-6}$ per epoch, and the batch size is 256 samples per batch.

\section{Additional results}

\subsection*{Latent space structure.}
\label{app:ls}
\begin{figure}[]
\begin{subfigure}{.5\textwidth}
  \centering
  \includegraphics[width=.9\textwidth]{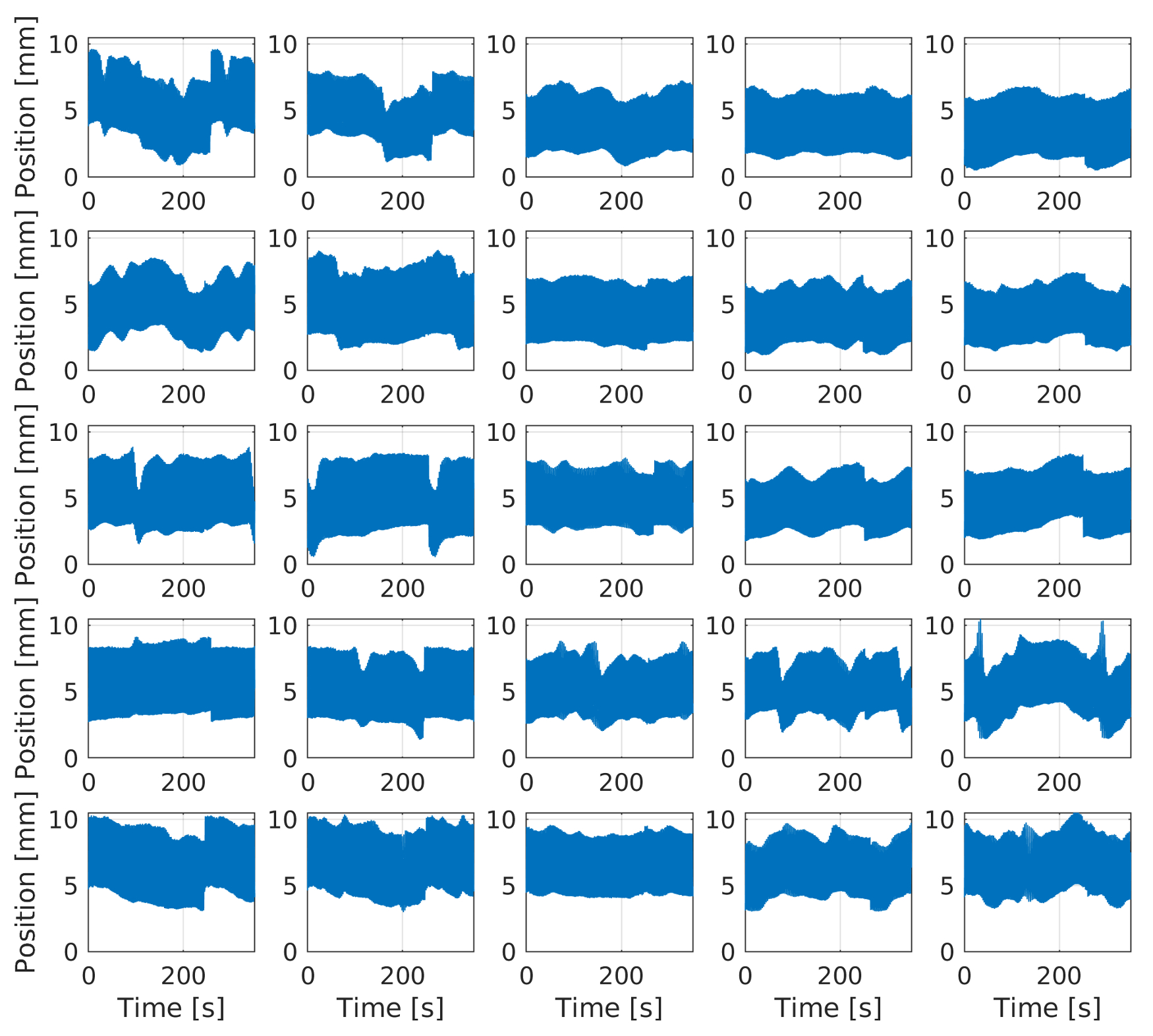}    
  \caption{AAE 2D latent space samples.}
  \label{fig:aaels}
\end{subfigure}
\begin{subfigure}{.5\textwidth}
  \centering 
  \includegraphics[width=.9\textwidth]{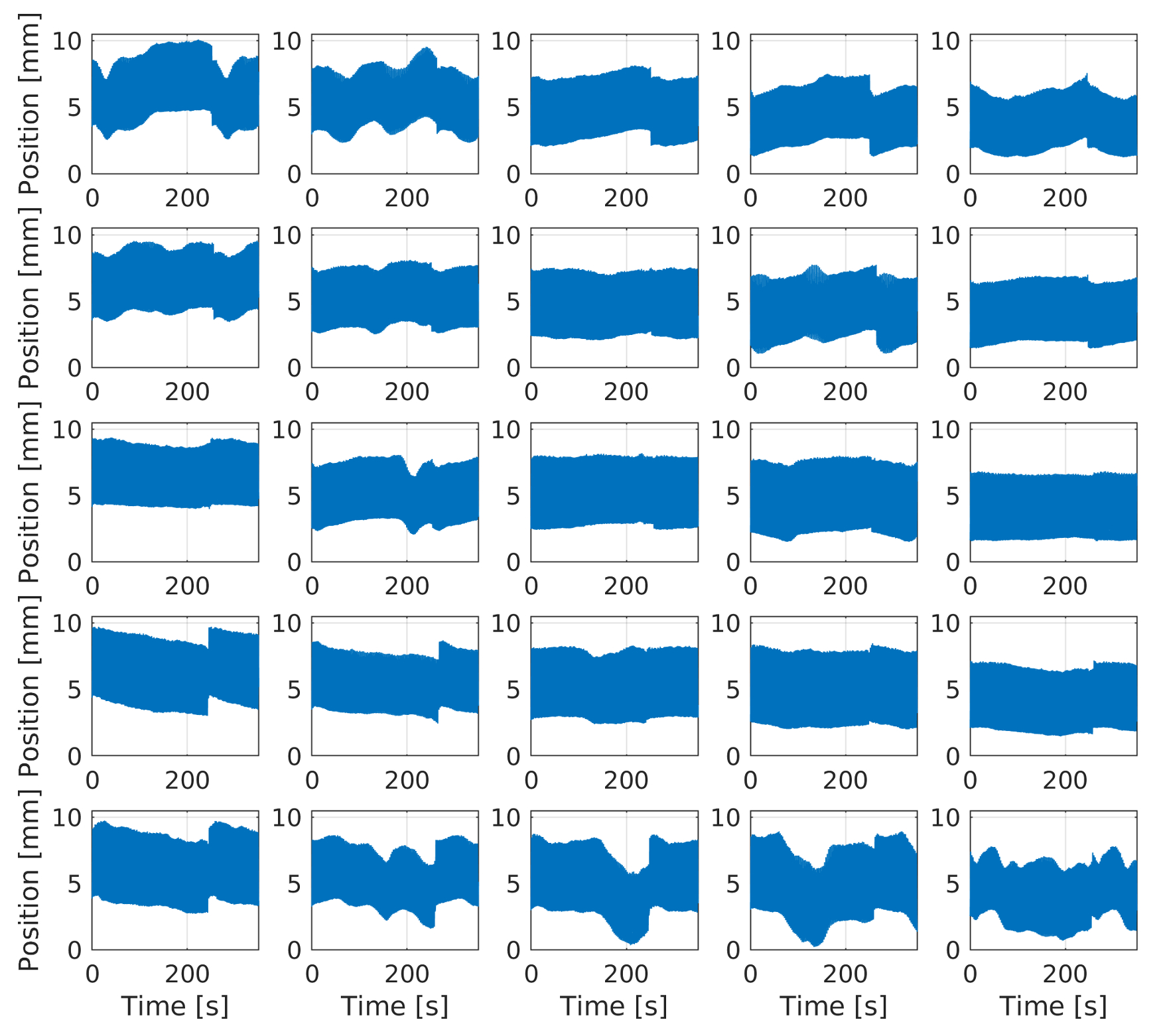} 
  \caption{VAE 2D latent space samples.}
  \label{fig:vaels}
\end{subfigure}
\begin{subfigure}{.5\textwidth}
  \centering
  \includegraphics[width=.9\textwidth]{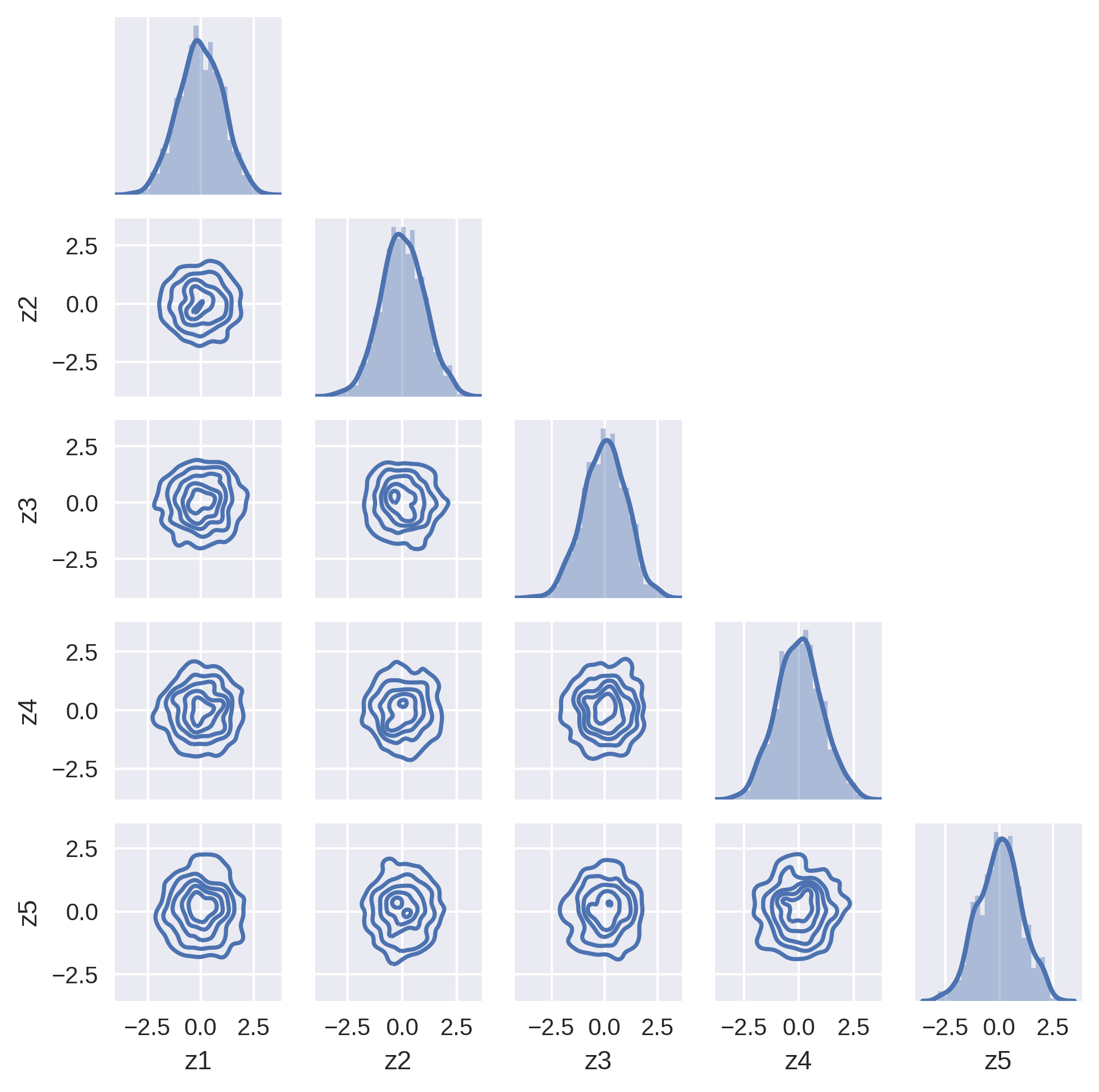}   
  \caption{AAE encodings distribution.}	
  \label{fig:lsaae}
\end{subfigure}
\begin{subfigure}{.5\textwidth}
  \centering 
  \includegraphics[width=.9\textwidth]{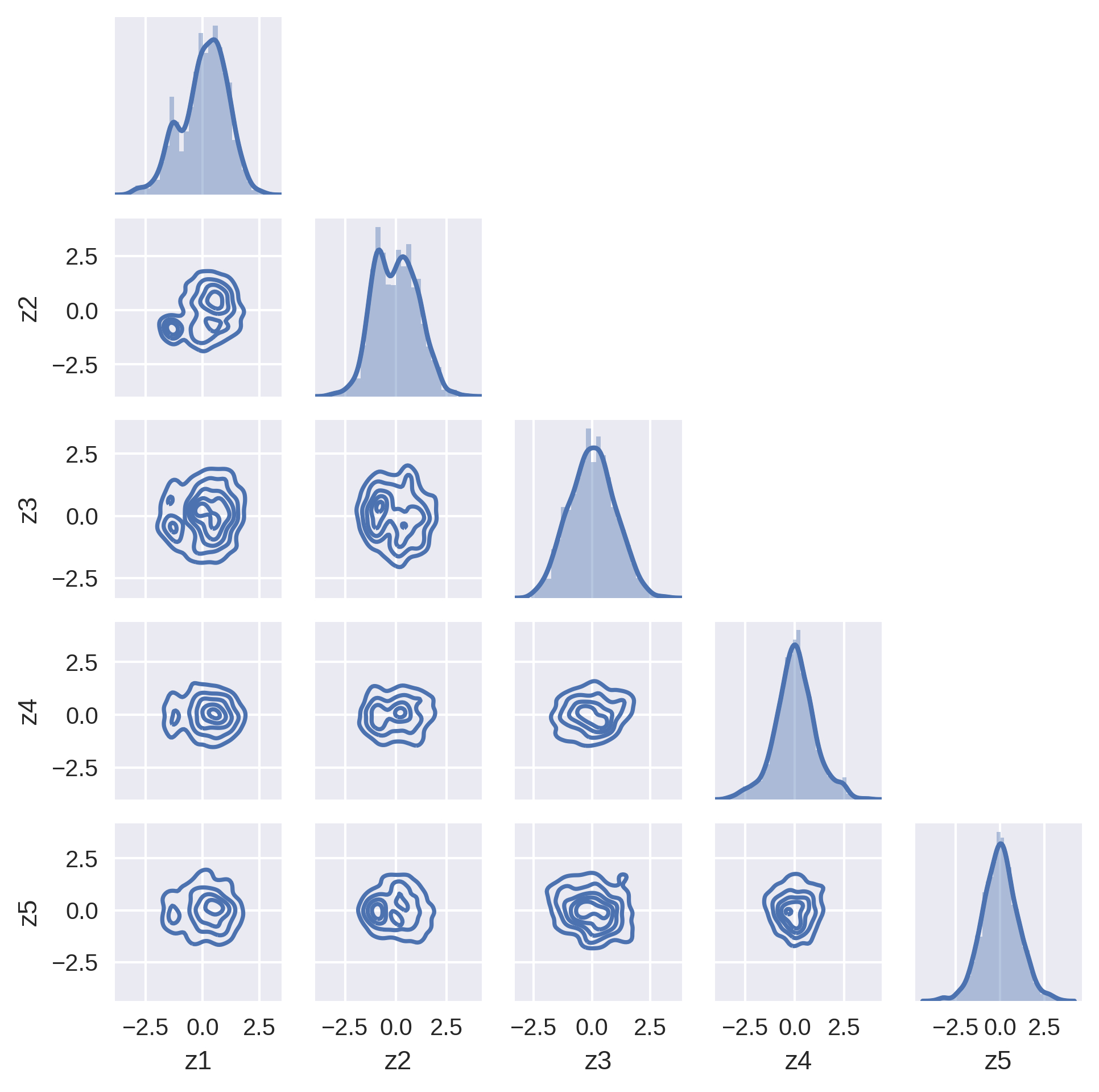}
  \caption{VAE encodings distribution.}		
  \label{fig:lsvae}
\end{subfigure}
\caption{(a, b) Sampled signals corresponding to a grid of evenly spaced encodings in a two-dimensional latent space. The grid consists of 25 equally spaced points covering the squared region with corner coordinates (-1.5,-1.5), (-1.5,1.5), (1.5,-1.5), (1.5,1.5). (c,d) Distribution of the dataset encodings in a five-dimensional latent space.}
\label{fig:latents}
\end{figure}

To visualize how the latent space is structured, we train the AAE and VAE frameworks using a two-dimensional latent space. \figref{aaels} and \figref{vaels} show samples from a grid of equally spaced $\bm{z}$ in such latent space. The square grid is defined as 25 equally spaced points between [-1.5, 1.5] in each of the two axis. According to the prior Gaussian distribution on the latent space, the samples in the center are more likely to be observed than the ones at the corners. Signals from nearby regions in the latent space show similar traits, such as the same type of irregularities or similar amplitudes and exhale positions.

To visualize any possible mismatch between the Gaussian prior and the aggregated posterior in the latent space, we plot the distribution of the encodings of all points in the dataset (i.e. the approximated aggregated posterior distribution). \figref{lsaae} and \figref{lsvae} show the distribution over a five-dimensional latent space for the AAE and VAE, respectively. The encodings of the AAE encoder are closely distributed to the prior Gaussian distribution.

\subsection*{Semi-supervised population model.} 
\begin{figure}[]
	\centering 
	\includegraphics[width=\textwidth]{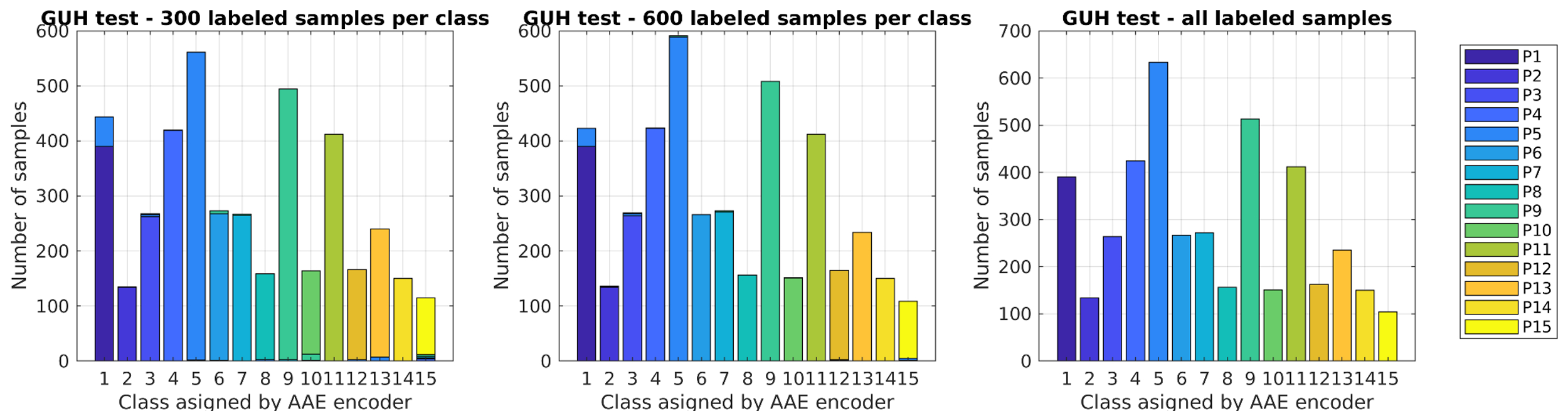}
	\caption{Performance of the population SAAE classification head on the test GUH data, when using 300 and 600 labels per class, and all the available labeled samples during the supervised training step. The abscissa displays the label assigned by the encoder, while the legend shows the color code for the true labels. The color of each of the bars shows the true label of the samples assigned to a certain class by the encoder.}
	\label{fig:popt}
\end{figure}

Using the SAAE framework, it is possible to train a population model that classifies and generates data from all the patients in the GUH dataset. The encoder classifies each signal into 15 classes corresponding to each of the patients in the dataset. The models are trained using a 80\%-10\%-10\% train, validation and test set split.  The population model can be used to classify and assign a new breathing sample to the most similar patient, and subsequently generate breathing samples from such patient. \figref{popt} shows the classification performance using 300 and 600 labeled data points per class during the supervised classification training phase, which corresponds to approximately 12.5\% and 25\% of the labels in the dataset. For comparison, we plot the performance when the labels of all the data points are used during training. The dimensionality of the latent space and the classification head is set to 15 ($C=15$, $N=15$).

One of the main limitiations when training patient-specific models is the size of the dataset. Deep learning methods are data-driven and require a significant amount of different examples to achieve good generalization. The GUH dataset is formed by long breathing signals (in some cases multiple signals per patient) from which between 1200 and 5000 samples can be obtained for each patient. This is not generally the case for the data recorded in clinics on a regular basis, usually consisting of short breathing signals of few minutes, as is the case for the majority of the EMC dataset. This highlights the need for population models in the specific case of breathing.

\subsection*{Sampling the semi-supervised models.}
\label{app:ss}
\begin{figure}[]
\begin{subfigure}{\textwidth}
	\centering
	\includegraphics[width=.8\textwidth]{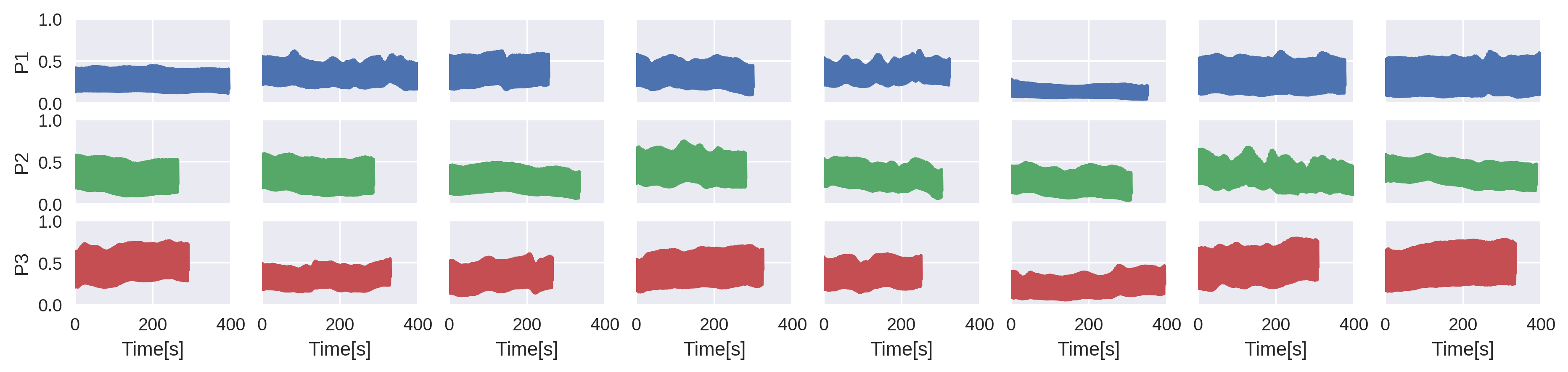} 
	\caption{Random samples from the baseline shift model. Rows 1, 2 and 3 visibly show signals with regular, downward and upwards baseline shifts.}		
	\label{fig:bsgen}
\end{subfigure}

\begin{subfigure}{\textwidth}
	\centering
	\includegraphics[width=.8\textwidth]{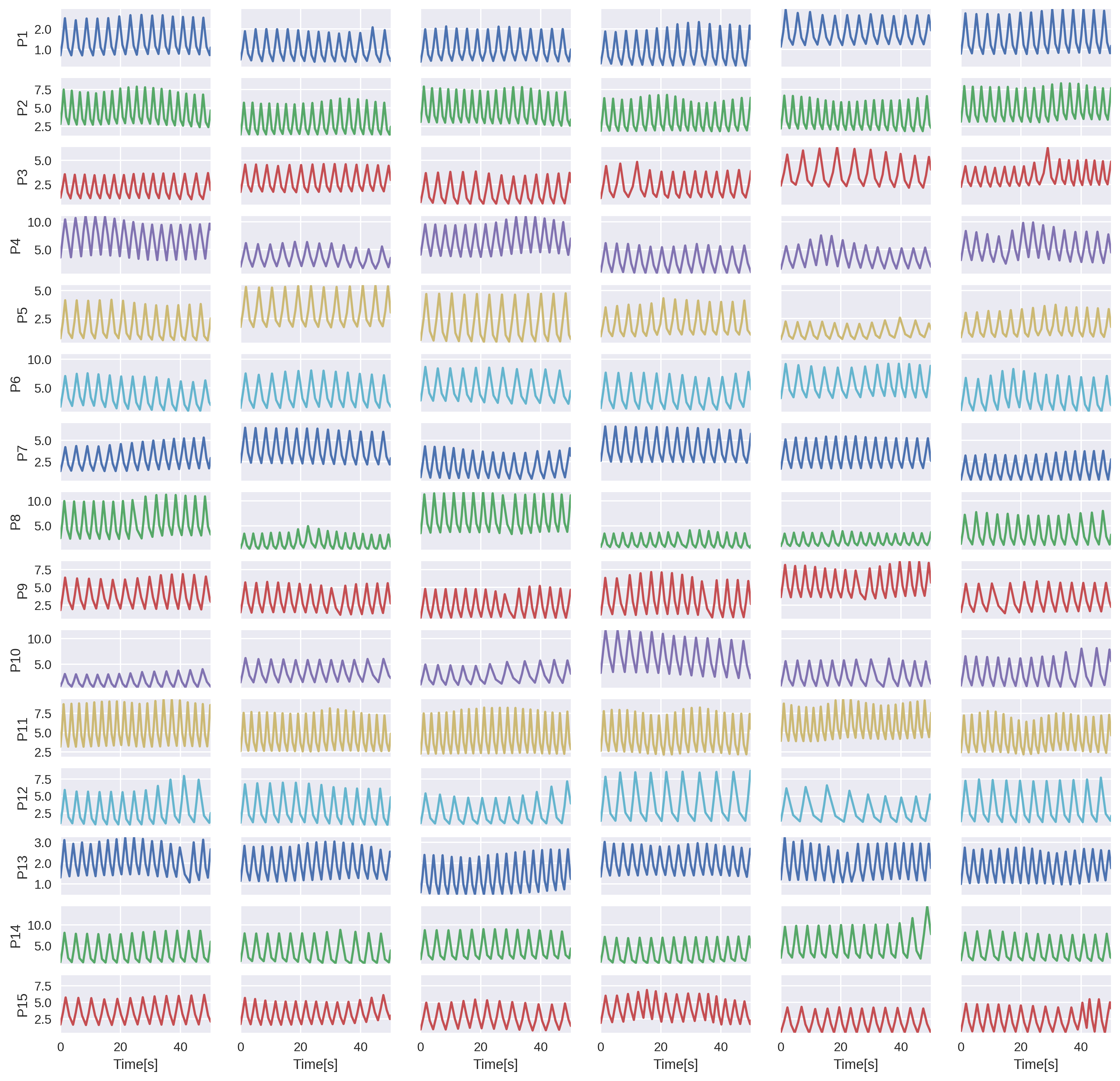} 
	\caption{Random samples from the population model. Each of the rows represents a patient from the dataset.}		
	\label{fig:popgen}
\end{subfigure}
\caption{Randomly generated signals for each class $\bm{y}$ in the baseline shifts and the population model. Each row represents (a) a type of baseline shift --- regular (C1), downwards baseline shifts (C2) and upwards baseline shifts (C3) --- or (b) the patient in a cohort. For each class, different $\bm{z}$ values are independently sampled from the isotropic Gaussian distribution $\mathcal{N}(\bm{z};\bm{0},\bm{I})$. Note that amplitudes can sometimes notably differ (P8) and periods are usually similar within each patient (P11).}
\end{figure}

The SAAE models can generate breathing signals that present a certain type of irregularity or resemble breathing from a certain patient. First, a class $\bm{y}$ is obtained from the encoder or sampled from the categorical prior, and then the Gaussian sub-manifold representing breathing of that particular class is sampled according to the prior distribution $p(\bm{z})$. \figref{bsgen} displays samples for each of the three classes in the baseline shift model trained using 12\% of labels in the dataset, while \figref{popgen} shows samples from each patient in the population based model that is trained using 600 labeled examples per class.

\end{document}